\documentclass[a4paper,fleqn,colorlinks,hidelinks]{cas-sc}

\usepackage[utf8]{inputenc}
\usepackage[T1]{fontenc}
\usepackage{microtype}
\usepackage{amsmath,amssymb}
\usepackage{graphicx}
\usepackage{booktabs}
\usepackage{multirow}
\usepackage{array}
\usepackage{tabularx}
\usepackage{longtable}
\usepackage{pdflscape}
\usepackage{makecell}
\usepackage{xcolor}
\usepackage{enumitem}
\usepackage{subcaption}

\usepackage{xurl}

\usepackage[section]{placeins}

\usepackage{algorithm}
\usepackage{algpseudocode}

\usepackage[authoryear,round]{natbib}

\graphicspath{{figures/}}

\begin{document}
\let\WriteBookmarks\relax

% Archival-preprint footer: name the target journal explicitly.
\ExplSyntaxOn
\cs_set:Npn \__first_footerline:
{
  \group_begin:
  \small
  \sffamily
  \ifnum\theblind>0\relax
  \else
    \__short_authors: :~
  \fi
  { \rmfamily \itshape Preprint }
  \group_end:
}
\ExplSyntaxOff

\shorttitle{Energy Vision--Language--Action}
\shortauthors{Saad Saoud et al.}

\title[mode=title]{Energy Vision--Language--Action: A Controlled Multimodal Benchmark for Intent-Conditioned Residential Energy Management}

\author[1]{Lyes {Saad Saoud}}
\cormark[1]
\author[2]{Oualid Doukhi}
\author[3]{Ehsan Reihani}
\author[4]{Saeed Sepasi}
\author[2]{Deok Jin Lee}
\author[5]{Moussa Ayyash}
\author[6]{Reza Ghorbani}

\affiliation[1]{
    Mohamed bin Zayed University of Artificial Intelligence,
    Abu Dhabi,
    UAE
}

\affiliation[2]{
    Jeonbuk National University, Deokjin-gu, Jeonju-si, Jeollabuk-do, South Korea
}
\affiliation[3]{
     California State University, Bakersfield, USA
}
\affiliation[4]{
    Hawaii Natural Energy Institute, 
    Hawaii, HI,  USA
}
\affiliation[5]{
    Chicago State University,
Chicago, IL, USA
}
\affiliation[6]{ 
University of Hawai`i at Manoa,
Honolulu, HI, USA
}

\cortext[1]{Corresponding author. Lyes Saad Saoud}

\begin{abstract}
Vision--Language--Action (VLA) models are studied mainly in robotics, where visual observations and language instructions are mapped to physical actions. This paper introduces Energy Vision--Language--Action (EVLA), a controlled multimodal benchmark for intent-conditioned residential energy management. EVLA frames residential battery scheduling as a multimodal trajectory-prediction problem in which a synthetic red--green--blue (RGB) energy-field representation, a numerical operating state, and a natural-language objective are mapped to a 16-step battery-action trajectory generated by a finite-horizon sampling-based reference trajectory generator. Source operating windows are derived from public residential electrical-load data, while electricity price, battery state of charge, indoor temperature, and time of day are generated benchmark metadata. A hidden operating regime is encoded only through the energy-field texture, enabling paired changes in visual context while the explicit numerical state is held fixed. Crossing 439,203 retained base windows with three hidden regimes and five language objectives yields 6,588,045 multimodal instances. An initial study evaluates 36 named configurations over three training seeds using fixed subsets of 5,000 training, 500 validation, and 500 test instances. In the MobileNet-family modality comparison, removing processed language increases trajectory mean-squared error from $0.3856 \pm 0.0039$ to $0.8628 \pm 0.0001$, whereas removing vision yields $0.3843 \pm 0.0013$, comparable to the full model. The study reveals a pronounced asymmetry in modality use within the implemented multimodal predictor: its processed-language pathway is strongly associated with prediction quality, while the current RGB pathway provides no aggregate error advantage. The reported results characterize the fixed pilot subset and executed protocol rather than full-benchmark training. EVLA provides a controlled setting for studying how semantic intent and latent context influence residential energy-action prediction.
\end{abstract}

\maketitle

% \begin{center}
% \small\textit{Preprint submitted to Engineering Applications of Artificial Intelligence}
% \end{center}
\vspace{0.5em}

\section{Introduction}
\label{sec:introduction}

Residential electricity systems are becoming increasingly active and flexible. Behind-the-meter batteries, distributed photovoltaic generation, heat pumps, electric vehicles, dynamic tariffs, and demand-response programs allow households to adjust when and how electricity is consumed, stored, or supplied to the grid \citep{iea2022der,iea2024batteries,iea2024renewables,iea2024ev}. As these technologies become more widely deployed, residential energy-management systems must account for interacting factors such as electricity price, battery state of charge, indoor temperature, time of day, uncertain operating conditions, and user preferences. A resident may seek to reduce electricity costs during one period, preserve indoor comfort during another, maintain battery reserve for an outage, or contribute to grid-support objectives.

Residential energy management is therefore more than a forecasting problem. It is a constrained sequential decision problem in which actions must respond to the current operating condition while reflecting the objective specified by the user. In particular, the same measured residential state can require different battery actions under different user priorities. This creates a need for decision models that can combine numerical measurements, contextual information, and semantic user intent.

\begin{figure}
    \centering
    \includegraphics[width=0.98\textwidth]{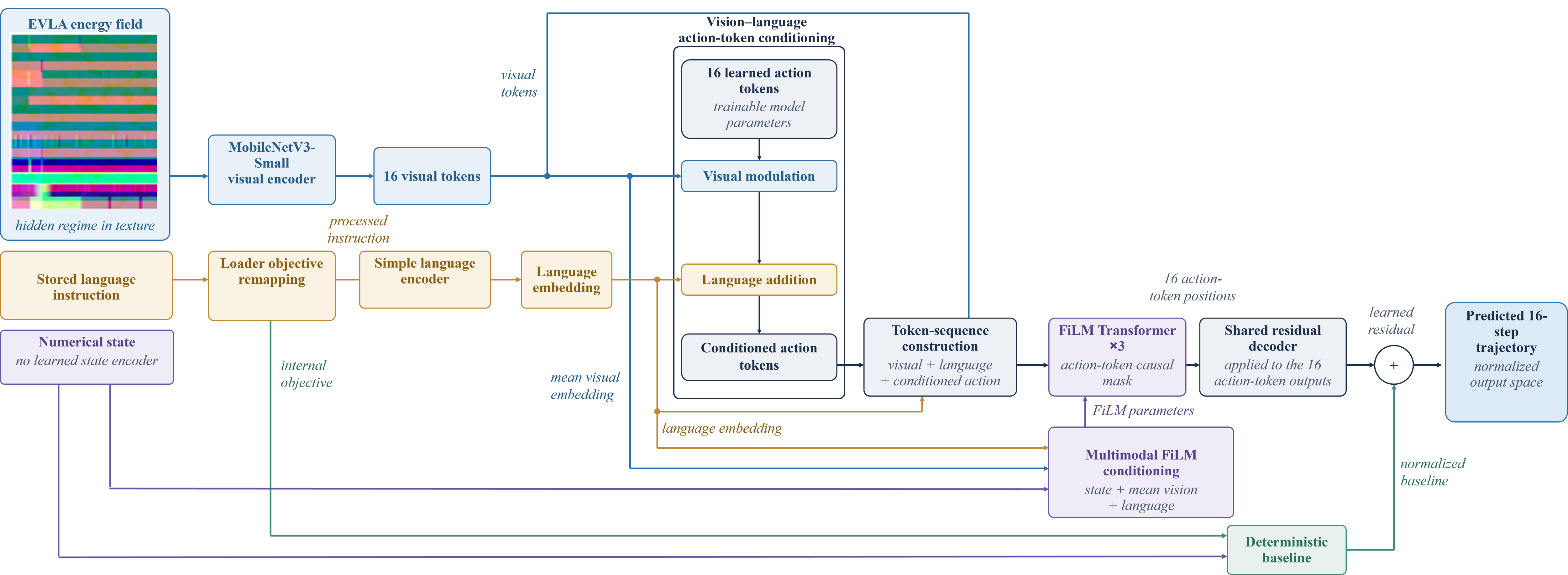}
    \caption{EVLA task formulation for residential energy management. A synthetic RGB energy-field observation, a three-variable numerical benchmark state, and a natural-language operating objective are mapped to a 16-step normalized battery-action trajectory. The controlled hidden regime is available to the predictor only through the texture of the energy-field representation.}
    \label{fig:evla_overview}
\end{figure}

\subsection{Motivation and Research Gap}

Residential energy-management systems traditionally rely on optimization-based controllers that explicitly represent forecasts, storage limits, comfort requirements, electricity tariffs, and system dynamics \citep{drgona2020mpcbuildings,michailidis2025mpc}. Learning-based methods can reduce online computation and improve adaptation, but most existing formulations still map numerical state variables to actions under a fixed objective \citep{yu2019drlsmarthome,laton2024drlhems,wang2025safedrl}. They therefore provide limited support for situations in which a resident changes the operating objective through natural language.

Language-based home-energy interfaces have begun to translate user requests into structured controller parameters \citep{michelon2025llmhems}. More recent studies use large language models for dynamic preference elicitation, agentic appliance scheduling, sustained multi-agent assistance, and iterative preference-conditioned reinforcement learning \citep{zhang2026llmhem,elmakroum2026agentic,jung2026hema,du2026adaptive}. These systems demonstrate growing interest in natural-language interaction for energy management; however, translating instructions into parameters, rules, or scheduling workflows differs from learning a controlled relationship among numerical benchmark states, contextual observations, processed language, and continuous battery-action trajectories.

VLA learning offers a useful abstraction for this problem because it combines visual observations and language instructions with action prediction \citep{brohan2023rt1,walke2023bridgedata,openx2024,octo2024,kim2025openvla}. Existing VLA benchmarks nevertheless focus primarily on robotics, autonomous driving, or graphical-interface interaction \citep{arai2025covla,lin2025showui}. Their observations and actions differ from those required for residential battery management.

A further limitation is that naturally collected energy variables are often correlated. A model may appear to use multiple modalities while relying primarily on one dominant input. Aggregate prediction error alone may therefore be insufficient to determine whether a model responds to language intent or contextual information.

The resulting research gap is the absence of a controlled benchmark that jointly relates numerical residential-state information, structured contextual observations, natural-language objectives, and continuous battery-control trajectories. EVLA addresses this gap by holding the source operating condition fixed while independently varying the language objective and a hidden contextual regime represented through an RGB energy-field channel.

\subsection{EVLA Formulation}

Each EVLA instance contains three observable inputs: an RGB energy-field representation, numerical residential-state information, and a natural-language operating objective. The target is a continuous battery-control trajectory over a 16-step prediction horizon, as illustrated in Fig.~\ref{fig:evla_overview}.

The metadata associated with each residential operating window include electricity price, battery state of charge, indoor temperature, and hour of day. The learned state encoder receives normalized electricity price, battery state of charge, and indoor temperature. Hour of day remains metadata and is used when constructing the solar-generation proxy employed by the operational diagnostics, but it is not supplied to the state encoder.

The language input \(L\) represents one of five operating-objective classes: energy saving, comfort, balanced operation, grid support, or eco operation. Each class is associated with multiple natural-language instructions. The visual input is a synthetic RGB energy-field representation derived from the same residential operating window. A hidden operating regime \(Z\) is excluded from the numerical state and encoded through the spatial texture of this representation.

For each tuple \((S,Z,L)\), a finite-horizon sampling-based reference trajectory generator produces a 16-step battery-action trajectory,

\begin{equation}
    A^\star = f_\star(S,Z,L),
    \label{eq:benchmark_relation_intro}
\end{equation}

where \(A^\star\) is the selected trajectory among sampled candidates evaluated under action bounds, battery-state limits, smoothness penalties, objective-dependent costs, and regime-dependent state evolution.

The learning task is to approximate \(A^\star\) from the observable inputs. Because \(Z\) is available only through the image, sensitivity to the hidden regime must be learned from the energy-field representation. Likewise, independent variation of \(L\) prevents the assigned objective from being inferred reliably from the numerical state alone.

The paired factorial organization is central to EVLA. For every retained base operating window, the source measurements and generated physical metadata are fixed while \(Z\) and \(L\) are varied at the benchmark level. The same residential condition can therefore be examined under several contextual regimes and user objectives. Importantly, \(Z\) is not a pure image-only intervention in the target generator: it also changes the candidate-action distribution and heuristic rollout dynamics. The paired construction therefore supports controlled regime- and language-conditioned comparisons, while stronger causal isolation requires the additional controls discussed in Sections~\ref{sec:discussion} and~\ref{sec:limitations}.
\subsection{Illustrative Residential Application Scenario}
\label{subsec:application_scenario}

Consider a household energy-management system with access to battery state of charge, indoor temperature, electricity price, and a recent window of household energy measurements. Under the same measured operating condition, the resident may request different priorities. An energy-saving instruction favors cost-aware battery operation, a comfort-oriented instruction emphasizes indoor conditions, and a grid-support instruction introduces a different operational objective. The corresponding battery trajectories may therefore differ even though the measured condition remains fixed.

EVLA reproduces this situation by preserving the base operating window while varying the language objective and hidden regime. The resulting paired instances make it possible to ask whether a predictor changes its trajectory when the requested priority changes and whether it responds to contextual information available only through the energy-field representation. This application scenario motivates the benchmark design; the assumptions that separate the generated targets from physical closed-loop deployment are addressed explicitly in Section~\ref{sec:limitations}.
\subsection{Contributions}

The study makes five contributions:

\begin{enumerate}[leftmargin=*]

    \item \textbf{A VLA-style formulation for residential energy management.}
    EVLA combines a synthetic RGB energy-field observation, numerical benchmark-state information, and a natural-language operating objective to predict a continuous 16-step battery-action trajectory. To the best of our knowledge, it is the first controlled VLA-style benchmark to combine these three inputs with continuous battery-trajectory targets under paired counterfactual variation for residential energy management.

    \item \textbf{A paired factorial dataset design.}
    EVLA is organized around \(A^\star=f_\star(S,Z,L)\). Each retained base operating window is expanded across three hidden regimes and five language objectives, producing 15 related instances that support paired analysis of state, visual context, and language intent.

    \item \textbf{A constraint-aware target generator.}
    Reference trajectories are produced by a finite-horizon sampling-based shooting procedure with battery-state limits, action bounds, smoothness terms, objective-dependent costs, comfort-related effects, and regime-dependent dynamics. The generator provides a consistent reference mechanism for the benchmark.

    \item \textbf{A large, structurally audited multimodal collection.}
    EVLA contains 439,203 retained base operating windows from 19 household files and 6,588,045 generated multimodal instances. The audit covers record structure, trajectory length, consistency of metadata across related instances, and sampled image availability.

    \item \textbf{A systematic multimodal evaluation.}
    Thirty-six named configurations are evaluated over three training seeds using fixed subsets of 5,000 training, 500 validation, and 500 test instances. The study includes modality ablations, visual-backbone comparisons, fusion and decoder variants, and neural baselines.

\end{enumerate}

The evaluation exposes an important asymmetry: processed language is strongly associated with prediction quality, whereas the current visual pathway contributes little to aggregate trajectory error. This gap between information encoded by the benchmark and information exploited by the predictor motivates the modality-specific analysis developed later in the paper.
\section{Related Work}
\label{sec:related_work}

EVLA lies at the intersection of residential energy management, 
language-conditioned human--system interaction, controlled multimodal 
learning, and Vision--Language--Action (VLA) modeling. Energy-management 
research has focused primarily on optimization, forecasting, and numerical 
control, whereas VLA research has concentrated mainly on robotics, driving, 
and digital interaction. This section reviews the closest areas of research 
and positions EVLA with respect to them.

\subsection{Optimization and Learning for Residential Energy Management}

Residential and building energy-management systems commonly employ model
predictive control (MPC), mixed-integer programming, stochastic optimization,
and robust optimization. These approaches allow physical constraints, comfort
limits, storage capacities, electricity tariffs, and operational costs to be
represented explicitly \citep{drgona2020mpcbuildings,michailidis2025mpc}.
MPC is particularly suitable for coordinating batteries, HVAC systems,
renewable generation, and electric-vehicle charging over a finite prediction
horizon.

Battery energy-storage systems have also been investigated for operational
objectives such as peak shaving and power smoothing under high renewable-energy
penetration. \citet{reihani2016peak} demonstrated the use of battery storage
for peak-load reduction and smoothing of a distribution-grid load profile,
illustrating the importance of combining forecasting with storage scheduling.

Data-driven forecasting provides another important component of residential
energy management. \citet{saoud2022household} combined the stationary wavelet
transform with Transformer models for multiresolution household energy-consumption prediction. More generally, time-series learning has been studied
using adaptive neural representations; for example, \citet{saoud2020octonion}
developed a metacognitive octonion-valued neural network and evaluated it on
several forecasting tasks, including household power consumption.
Renewable-energy forecasting has also been investigated using nonlinear neural
representations, including 24-hour-ahead global solar-irradiation forecasting
\citep{saoud2018solar}.

These forecasting and storage-management studies provide useful foundations
for learning temporal energy-system behavior. EVLA addresses a different but
complementary problem: rather than forecasting a single energy variable or
optimizing storage under a fixed objective, it studies whether a learned
predictor can map numerical residential state, contextual information, and a
natural-language operating objective to a continuous battery-action trajectory.

Deploying optimization-based controllers across heterogeneous households can,
however, require household-specific models, forecasts, solver settings, and
objective weights. Changes in resident preferences may also require the
controller objective to be modified or retuned. Learning-based approaches have
therefore been investigated as a means of reducing online computation and
improving adaptation. Deep reinforcement learning has been applied to HVAC
operation, battery scheduling, renewable-energy utilization, and
electricity-cost reduction \citep{yu2019drlsmarthome,laton2024drlhems}.
Safe and constrained learning methods have additionally combined learned
policies with explicit constraint handling or MPC-based safeguards
\citep{wang2025safedrl}.

Most of these formulations map numerical measurements to actions under a fixed
reward or predefined operating objective. EVLA complements these approaches by
using the operating objective as a language input and learning a continuous
battery-action trajectory against reference trajectories produced by a
constraint-aware sampling-based reference trajectory generator.

\subsection{Human-Interactive and Language-Conditioned Energy Management}

Conventional home energy-management systems generally require residents to 
express preferences through schedules, setpoints, operating modes, numerical 
weights, or predefined rules. These mechanisms do not correspond directly to 
the way residents naturally describe objectives such as reducing electricity 
cost, preserving comfort, limiting battery cycling, or supporting the grid.

Large-language-model interfaces for home energy management have begun to 
address this mismatch by translating user requests into structured controller 
settings \citep{michelon2025llmhems}. In such systems, language generally acts 
as an interface to an existing controller: the request is converted into a 
command, schedule, or objective parameter before numerical optimization or 
control is performed.

EVLA studies a related but distinct problem. Language is included directly in 
the trajectory-prediction task and is varied independently of the underlying 
operating window. This construction makes it possible to test whether changing 
the stated objective produces a corresponding change in the predicted 
battery-control trajectory.

\subsection{Controlled Multimodal Learning in Cyber-Physical Systems}

Decision making in cyber-physical systems can depend on heterogeneous inputs, 
including numerical measurements, images, text, forecasts, and contextual 
metadata. Merely supplying several modalities does not establish that a model 
uses each of them meaningfully. When inputs are correlated, a predictor may 
obtain low aggregate error by relying primarily on one dominant modality.

Controlled and paired comparisons help reveal such behavior. By holding 
an operating condition fixed while changing one input factor, the response of 
the model to that factor can be examined more directly. This principle is 
related to causal and counterfactual analysis, in which interventions are used 
to distinguish effects from observational associations 
\citep{pearl2009causality,peters2017elements}.

EVLA applies this principle by independently varying the hidden contextual 
regime \(Z\) and language objective \(L\) for each retained base operating 
window. The evaluated numerical state contains electricity price, battery 
state of charge, and indoor temperature, while additional temporal information 
is retained as metadata used during data construction. Consequently, samples 
can share the same numerical control inputs while differing in visual context 
or language intent.

This design distinguishes two questions. First, does a modality contain
information that changes the reference action? Second, does the learned model
respond to that information? EVLA examines both questions by combining
aggregate prediction error with controlled data construction, modality
ablations, regime-conditioned statistics, and sensitivity diagnostics.

\subsection{Vision--Language--Action Models}

VLA models have developed mainly in robotics, where visual observations and
language instructions are mapped to physical actions. RT-1 demonstrated robot
control from images and task instructions \citep{brohan2023rt1}, while
BridgeData V2 and Open X-Embodiment increased the scale and diversity of
cross-task and cross-robot training data
\citep{walke2023bridgedata,openx2024}. Generalist policies such as Octo and
OpenVLA have further investigated training and adaptation across diverse
robot-manipulation datasets \citep{octo2024,kim2025openvla}.

Recent systematic analysis has further emphasized multimodal-fusion strategies,
cross-modal alignment, action decoding, and the interaction among vision,
language, and control representations in VLA systems
\citep{din2026vlafusion}. These considerations are particularly relevant to
EVLA because its learning problem similarly requires heterogeneous modalities
to influence a continuous action representation, although the application
domain and action space differ substantially from robotic manipulation.

Existing VLA benchmarks remain centered primarily on manipulation, navigation,
locomotion, and related embodied action spaces
\citep{brohan2023rt1,octo2024,kim2025openvla}. The abstraction has also been
extended to other domains. CoVLA combines driving observations and language
descriptions with vehicle trajectories \citep{arai2025covla}, whereas ShowUI
maps interface screenshots and language instructions to graphical-interface
actions \citep{lin2025showui}.

Residential energy management requires a different formulation. Its action is
a continuous battery-power trajectory rather than a robot motion, vehicle
trajectory, or interface command. The action must also reflect battery limits,
tariffs, operating objectives, and temporal energy-system behavior. Moreover,
EVLA uses a structured RGB energy-field representation derived from residential
time-series data rather than a natural camera observation. This representation
encodes a controlled hidden regime whose contribution can be examined through
paired comparisons.

EVLA therefore adopts the general VLA relationship among observation, semantic
intent, and continuous action, but reformulates it for residential battery
management. Its reference actions are generated by a sampling-based reference trajectory generator rather than obtained from robotic demonstrations.

EVLA brings these research threads into a single trajectory-prediction setting for residential battery scheduling. Its distinguishing feature is not simply the use of several modalities, but the paired variation of language intent and visually encoded latent context for the same base operating window. That construction supports direct tests of whether a predictor responds to each modality instead of relying only on aggregate prediction accuracy.

\section{EVLA Dataset and Target Generation}
\label{sec:dataset}

EVLA is a controlled multimodal benchmark for residential energy-management
decision learning. Each generated sample contains a visual observation
\(V_{i,z}\), stored operating-state metadata \(S_i^{\mathrm{meta}}\), a
language objective \(L_\ell\), a hidden regime \(Z_i\), and a 16-step reference
battery-action trajectory \(A^\star_{i,z,\ell}\). The visual regime is encoded
through the energy-field image, whereas the language objective is supplied as
text. The reference trajectory is generated by a finite-horizon
sampling-based reference trajectory generator conditioned on the numerical state, hidden regime, and
language objective.

\subsection{Overview and Dataset Scale}

EVLA is derived from processed residential electrical-load records built from
19 REFIT households (Houses~1--13 and Houses~15--20)
\citep{murray2017refit,murray2016refitcleaned}. After structural validation
and window-bound checking, \(439{,}203\) residential operating windows are
retained as base instances. Each base instance is expanded across three hidden
regimes and five language objectives, yielding

\begin{equation}
N_{\mathrm{EVLA}}
=
439{,}203 \times 3 \times 5
=
6{,}588{,}045
\label{eq:dataset_size_clean}
\end{equation}

multimodal samples.

The hidden-regime set is
\(\mathcal{Z}=\{0,1,2\}\), corresponding to low, neutral, and high hidden
thermal conditions, respectively. The same index is used by the visual construction:
\(Z=0\) maps to coarse spatial texture, \(Z=1\) to medium texture, and
\(Z=2\) to fine texture. Thus, ``thermal condition'' names the designed hidden
regime, while coarse/medium/fine describes its image-space encoding; these labels
do not imply that texture itself is a measured thermal field. The language-objective set is
\(\mathcal{L}=\{\text{energy saving},\text{comfort},\text{balanced},
\text{grid support},\text{eco}\}\).

\begin{figure}
    \centering
    \includegraphics[width=0.98\textwidth]{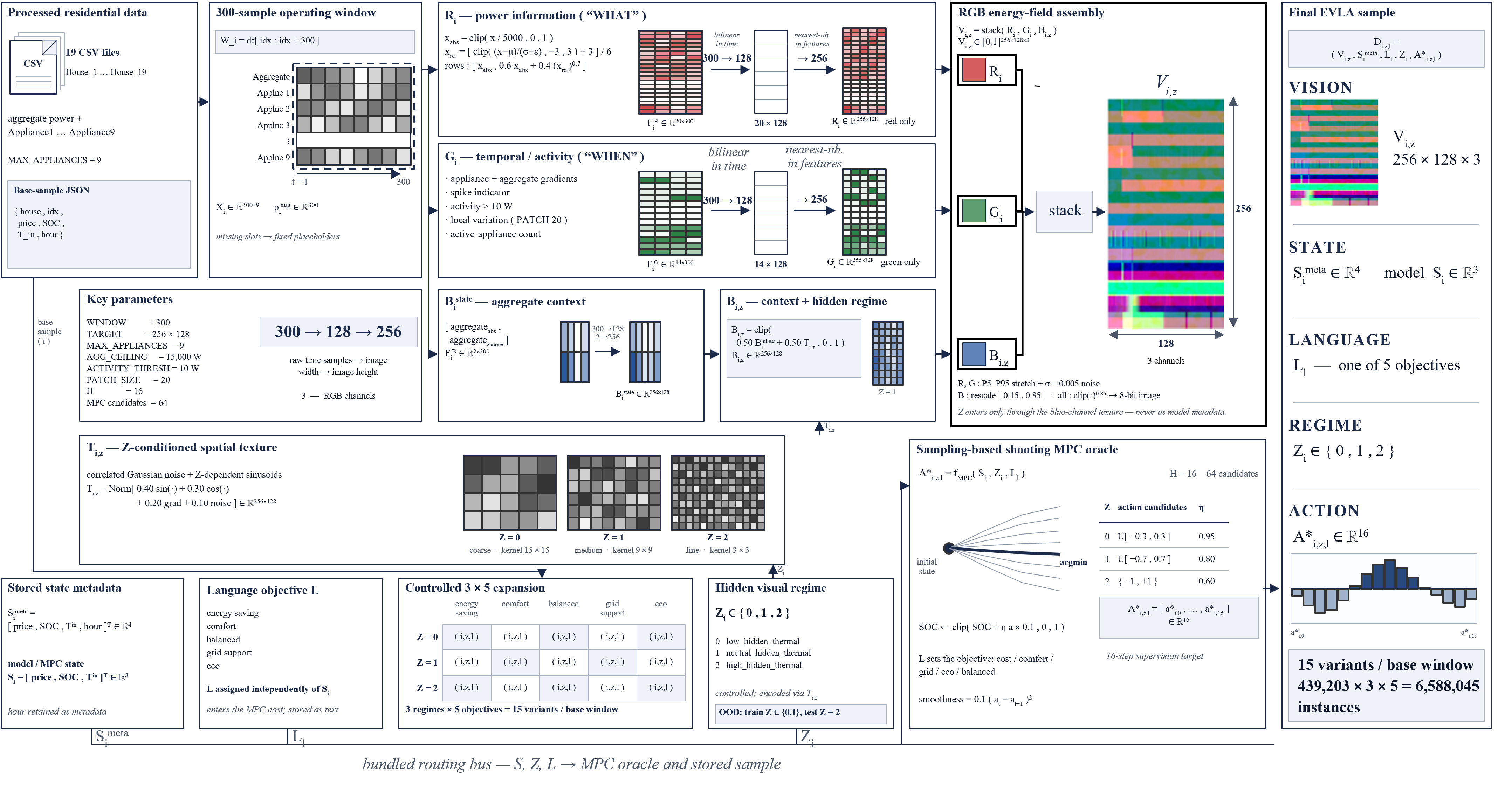}
    \caption{EVLA dataset- and target-generation pipeline. Processed residential
    CSV records and base-sample JSON metadata are used to extract
    300-sample operating windows. Each window is converted into an RGB
    energy-field representation of size \(256\times128\times3\), expanded
    across three hidden regimes and five language objectives, and paired with a
    16-step reference battery-action trajectory generated by a
    sampling-based shooting reference trajectory generator.}
    \label{fig:dataset_pipeline}
\end{figure}

\subsection{Operating-Window Extraction and Stored Metadata}

For each retained sample, the generator extracts a residential operating window

\begin{equation}
W_i = df[\mathrm{idx}_i:\mathrm{idx}_i+300],
\label{eq:window_clean}
\end{equation}

where the window length is fixed at 300 samples. The corresponding residential
measurements contain the aggregate active-power signal and up to
\(9\) appliance-level signals. If a house contains fewer than 9 appliances,
the missing appliance slots are filled with fixed placeholders during feature
construction so that all samples share the same channel dimensions.

Each retained window is also associated with stored operating-state metadata

\begin{equation}
S_i^{\mathrm{meta}}
=
[p_i,\; q_i,\; T_i^{\mathrm{in}},\; \tau_i]^\top \in \mathbb{R}^4,
\label{eq:state_meta_clean}
\end{equation}

where \(p_i\) is the electricity price, \(q_i\) is the battery state of
charge, \(T_i^{\mathrm{in}}\) is the indoor temperature, and \(\tau_i\) is
the hour of day. The predictor and the reference trajectory generator use the numerical state

\begin{equation}
S_i=[p_i,\; q_i,\; T_i^{\mathrm{in}}]^\top \in \mathbb{R}^3,
\label{eq:model_state_clean}
\end{equation}

while the hour is retained as stored metadata.

\subsection{RGB Energy-Field Construction}

Each 300-sample residential operating window is converted into a synthetic RGB
energy-field image of size \(256\times128\times3\). The red, green, and blue
channels encode complementary information derived from the same source window.

The red channel \(R_i\) represents \emph{power information}. It is constructed
from appliance-level and aggregate-load magnitude features using an absolute
scaling branch and a relative within-window normalization branch. With a
maximum of 9 appliances, this stage produces approximately
\(20\times300\) feature rows before resizing.

The green channel \(G_i\) represents \emph{temporal/activity information}. Its
feature rows include appliance and aggregate gradients, spike indicators,
activity indicators with a \(10\)~W threshold, local aggregate variation
computed with a patch size of 20, and the number of simultaneously active
appliances. This stage produces approximately \(14\times300\) feature rows
before resizing.

The blue pre-channel \(B_i^{\mathrm{state}}\) represents \emph{aggregate-load
context}. It is built from the absolute-scale and standardized aggregate-load
signals, yielding a \(2\times300\) feature matrix before resizing.

All feature matrices are converted to image format using the same procedure:
the temporal axis is first resized from 300 to 128 by bilinear interpolation,
and the feature axis is then resized to 256 by nearest-neighbor interpolation.
Thus, the final single-channel outputs satisfy

\[
R_i,\; G_i,\; B_i^{\mathrm{state}} \in \mathbb{R}^{256\times128}.
\]

The hidden regime \(Z_i\in\{0,1,2\}\) generates a regime-specific spatial
texture \(T_{i,z}\in\mathbb{R}^{256\times128}\). The three regimes correspond
to coarse, medium, and fine spatial structure, respectively. The final blue
channel is then formed as

\begin{equation}
B_{i,z}
=
\operatorname{clip}
\!\left(
0.50\,B_i^{\mathrm{state}}
+
0.50\,T_{i,z},
0,1
\right).
\label{eq:blue_channel_clean}
\end{equation}

The final RGB energy field is

\begin{equation}
V_{i,z}
=
\operatorname{stack}(R_i,G_i,B_{i,z})
\in [0,1]^{256\times128\times3}.
\label{eq:rgb_clean}
\end{equation}

After stacking, the generator applies a light post-processing step:
a 5th--95th percentile contrast stretch to the red and green channels, a
soft rescaling of the blue channel to \([0.15,0.85]\), a small
zero-mean Gaussian perturbation (\(\sigma=0.005\)) to the red and green
channels, and a final power transformation before 8-bit image quantization.

\subsection{Controlled Expansion over Regime and Language}

Each retained base operating window is expanded over the Cartesian product
\(\mathcal{Z}\times\mathcal{L}\). Hence, every base sample produces
\(3\times5=15\) controlled variants.

The language objective is one of five classes:
\emph{energy saving}, \emph{comfort}, \emph{balanced},
\emph{grid support}, and \emph{eco}. One paraphrase is selected from the
corresponding objective pool and stored as the language input \(L_\ell\).
Within the benchmark construction, language is assigned independently of the
physical operating state. For a fixed base window, changes in \(L_\ell\) alter
the textual objective and the target-generation cost, while changes in \(z\)
alter both the visual texture and the reference-generator dynamics.

\subsection{Sampling-Based Reference Trajectory Generation}

For every \((i,z,\ell)\) combination, EVLA generates a 16-step reference
battery-action trajectory using a finite-horizon sampling-based shooting MPC
reference trajectory generator. The generator takes the numerical state \(S_i\), the hidden regime
\(z\), and the language objective \(L_\ell\) as inputs, samples 64 candidate
action sequences, rolls them forward through heuristic state-update equations,
evaluates a language-dependent objective, and retains the lowest-cost sampled
sequence.

Formally, the generated target is

\begin{equation}
A^\star_{i,z,\ell}
=
f_{\mathrm{ref}}(S_i,z,L_\ell)
=
[a^\star_{i,0},\ldots,a^\star_{i,15}]^\top \in \mathbb{R}^{16}.
\label{eq:reference_generator_clean}
\end{equation}

The horizon is \(H=16\), and the action candidate set depends on the regime:
for \(z=0\), actions are sampled from \(\mathcal{U}[-0.3,0.3]\); for \(z=1\),
from \(\mathcal{U}[-0.7,0.7]\); and for \(z=2\), from \(\{-1,+1\}\). The
corresponding SOC-efficiency factors are \(0.95\), \(0.80\), and \(0.60\),
respectively. The rollout clips SOC to \([0,1]\) and uses a simple synthetic
temperature proxy whose drift and action coupling are also regime-dependent.

The language objective specifies the stage cost used by the reference trajectory generator:
energy saving emphasizes electricity cost, comfort emphasizes thermal
deviation, balanced combines cost and comfort, grid support favors
grid-oriented actions, and eco introduces a phase-dependent criterion. All
objectives also include an action-smoothness penalty
\(0.1(a_t-a_{t-1})^2\).

The resulting multimodal sample is therefore

\begin{equation}
D_{i,z,\ell}
=
\left(
V_{i,z},
S_i^{\mathrm{meta}},
L_\ell,
Z_i,
A^\star_{i,z,\ell}
\right).
\label{eq:final_sample_clean}
\end{equation}

\subsection{Benchmark Semantics}

EVLA deliberately uses a synthetic target generator so that state, language, and hidden regime can be varied under controlled conditions. The energy-field image is derived from residential power signals, the regime is a designed latent factor, and \(A^\star\) is the lowest-cost sampled trajectory under the specified candidate set and heuristic rollout equations. These choices make paired modality interventions possible and define precisely what the supervised target represents.

The resulting trajectories are benchmark references tied to this generator. Their physical interpretation depends on the simplified battery and temperature updates described above; the consequences of those assumptions, including action units and closed-loop validity, are treated in Section~\ref{sec:limitations}.
\section{EVLA Trajectory-Prediction Model}
\label{sec:controller}

The Energy--Vision--Language--Action (EVLA) predictor is trained to approximate
the reference trajectory-generation procedure introduced in
Section~\ref{sec:dataset}. For sample $i$, the learned model receives the RGB
energy-field observation $V_i$, a three-dimensional numerical state $S_i$, and
a processed language instruction $L_i$, and predicts the complete $H=16$ action
sequence in one forward pass. The model therefore implements
\begin{equation}
    \widehat{\widetilde{A}}_i
    =
    f_{\theta}(V_i,S_i,L_i)
    =
    \left[
        \hat{\widetilde a}_{i,0},\ldots,
        \hat{\widetilde a}_{i,15}
    \right]^{\top}
    \in\mathbb{R}^{16},
    \label{eq:evla_mapping}
\end{equation}
where $\theta$ contains all trainable parameters and the tilde denotes the
model's normalized action space. The common latent dimension is $d=256$.
The hidden benchmark regime $Z_i$ is \emph{not} passed to the predictor as a
numerical variable; any predictive effect of $Z_i$ must be recovered from the
regime-dependent texture already embedded in $V_i$.

Fig.~\ref{fig:evla_overview} gives the overall EVLA information flow. The
architecture has three learned information routes---vision, language, and
numerical state---plus a deterministic baseline route. Vision and language
condition 16 learned action tokens, the resulting tokens are concatenated with
the context tokens, three FiLM-conditioned Transformer blocks model the token
sequence, and a shared decoder predicts a bounded learned residual. The final
trajectory is obtained by adding this residual to a deterministic baseline.

\subsection{Target Representation and Action Scale}
\label{subsec:target_scale}

The executed learning code uses the fixed numerical scale
\begin{equation}
    P_{\mathrm{scale}}=5.
    \label{eq:pscale}
\end{equation}
For the stored reference trajectory $A_i^{\star}$, target position $k$ is mapped to
normalized model space as
\begin{equation}
    \widetilde a_{i,k}^{\star}
    =
    \operatorname{clip}\!\left(
        \frac{a_{i,k}^{\star}}{P_{\mathrm{scale}}},-1,1
    \right),
    \qquad k=0,\ldots,15.
    \label{eq:target_normalization}
\end{equation}
The reference trajectory generator itself produces benchmark actions within the regime-specific
candidate ranges stated in Section~\ref{sec:dataset}, whose largest magnitude is
1. Thus, \(P_{\mathrm{scale}}=5\) is an implementation-level learning/evaluation
scale rather than a physical bound established by the reference trajectory generator. Predictions and
targets are multiplied by \(P_{\mathrm{scale}}\) inside the executed loss and
metric code. The deterministic residual baseline is also constructed using this
same implementation scale and can therefore occupy a wider numerical range than
the sampled reference candidates.

Accordingly, the reported errors should be interpreted only in the evaluator's
internal action scale. They are not physical \(\mathrm{kW}\) errors, and direct
physical comparison between the reference-generator candidate bounds, the deterministic
baseline magnitude, and the SOC/energy diagnostics is not claimed. A future
release should use one action convention consistently from target generation
through training, evaluation, and physical interpretation.

\subsection{Input Representation and Preprocessing}
\label{subsec:model_inputs}

\begin{figure}
    \centering
    \includegraphics[width=0.96\textwidth]{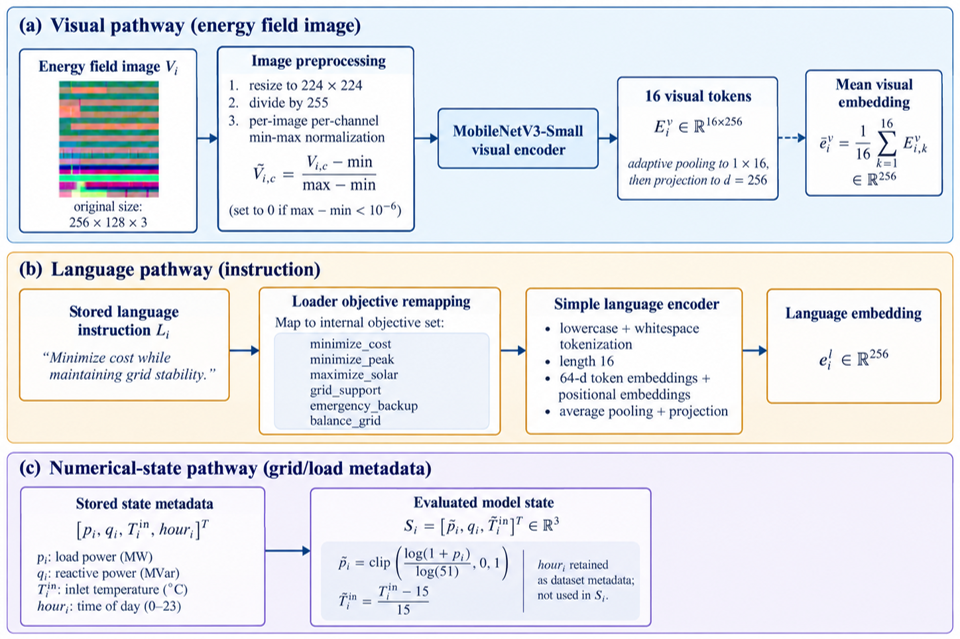}
    \caption{EVLA input preprocessing and modality encoders. (a) The
    $256\times128\times3$ energy field is resized to $224\times224$, normalized,
    and encoded into 16 visual tokens of width 256; their mean is retained for
    FiLM conditioning. (b) The stored instruction is mapped by the loader to an
    internal objective, processed by the simple language encoder, and projected
    to a 256-dimensional embedding. (c) The evaluated numerical state contains
    normalized price, SOC, and normalized indoor temperature; hour of day is
    retained as metadata and is not included in $S_i$.}
    \label{fig:evla_inputs}
\end{figure}

Figure~\ref{fig:evla_inputs} isolates the three model inputs and makes the
routing explicit. In particular, the language encoder receives only language;
the numerical state does not enter the text encoder. Conversely, the numerical
state is used directly by the FiLM context and deterministic baseline. The mean
visual embedding is used for FiLM conditioning (and for the auxiliary visual
head described below) but is not inserted as an additional token into the main
33-token sequence.

\subsubsection{Visual input}

Each generated energy-field image has native size
$256\times128\times3$. Before entering the visual encoder, it is resized to
$224\times224$ using bilinear interpolation and divided by 255. Each color
channel is then normalized independently within the image:
\begin{equation}
\widetilde V_{i,c}=
\begin{cases}
\dfrac{V_{i,c}-m_{i,c}}{M_{i,c}-m_{i,c}},
& M_{i,c}-m_{i,c}>10^{-6},\\[1mm]
0, & \text{otherwise},
\end{cases}
\label{eq:image_preprocessing}
\end{equation}
where $m_{i,c}=\min(V_{i,c})$, $M_{i,c}=\max(V_{i,c})$,
and $c\in\{R,G,B\}$.

No ImageNet mean--standard-deviation normalization is applied by this loader.
This is the preprocessing used in the reported experiments, including models
initialized from ImageNet weights; therefore, the pretrained backbones are not
evaluated under their canonical ImageNet input normalization. In addition, the
per-image, per-channel min--max transform in Eq.~\eqref{eq:image_preprocessing}
preserves within-image spatial structure but can suppress information carried by
absolute channel minima, maxima, and inter-image amplitude differences. No
preprocessing ablation was performed in the present study, so the visual-pathway
results should be interpreted conditional on this executed preprocessing choice.
The reported training pipeline does not use random crop, geometric
augmentation, or color augmentation; pixel shuffling is reserved for a
separate diagnostic experiment. If an image cannot be loaded, the loader uses
a zero $224\times224$ RGB tensor.

\subsubsection{Numerical state}

The predictor uses
\begin{equation}
    S_i
    =
    \left[
        \widetilde p_i,
        q_i,
        \widetilde T_i^{\mathrm{in}}
    \right]^{\top}
    \in\mathbb{R}^{3},
    \label{eq:model_state}
\end{equation}
with
\begin{equation}
    \widetilde p_i
    =
    \operatorname{clip}\!\left(
        \frac{\log(1+p_i)}{\log(51)},0,1
    \right),
    \qquad
    \widetilde T_i^{\mathrm{in}}
    =
    \frac{T_i^{\mathrm{in}}-15}{15}.
    \label{eq:state_normalization}
\end{equation}
The battery SOC $q_i$ is passed directly. The stored hour-of-day field remains
metadata and is not included in the evaluated state vector; consequently, no
cyclic hour encoding is used by the reported EVLA predictor.

\subsubsection{Trajectory input/target handling}

The first $H=16$ stored action values are used. A shorter trajectory, if
encountered, is padded by repeating its final value and a validity mask
$m_{i,k}$ records valid positions. The generated EVLA trajectories contain all
16 values. The first stored action is retained as $u_{i,0}$ and is used by the executed
training code as the reference preceding the first predicted position in the
smoothness term. Because this value is the first target action rather than a
measured pre-horizon control command, the \(k=0\) smoothness contribution is a
target-anchored implementation choice, not a physically observed action
transition.

\subsection{Visual Encoder}
\label{subsec:visual_encoder}

The main architecture uses MobileNetV3-Small as the visual encoder. After the
backbone feature extractor, adaptive pooling produces one spatial row with
$H=16$ positions. Each position is projected to the common model dimension,
producing
\begin{equation}
    E_i^v
    =
    \phi_v(\widetilde V_i)
    =
    \left[e_{i,0}^v,\ldots,e_{i,15}^v\right]
    \in\mathbb{R}^{16\times256}.
    \label{eq:visual_tokens}
\end{equation}
The projection uses a linear layer, layer normalization, GELU, and a second
linear layer. For MobileNetV3-Small, the 576-dimensional backbone output is
projected to 256 dimensions. The backbone is initialized from ImageNet weights
and is frozen except for its final three feature blocks.

The mean visual embedding is
\begin{equation}
    \overline e_i^v
    =
    \frac{1}{16}\sum_{k=0}^{15}e_{i,k}^v
    \in\mathbb{R}^{256}.
    \label{eq:mean_visual_embedding}
\end{equation}
It is used in the FiLM context and by the auxiliary vision head, but is not
inserted directly into the token sequence. The auxiliary head has dimensions
$256\rightarrow128\rightarrow 2H$ with ReLU and dropout and is trained against
the loader's load and solar-proxy sequences.

Backbone sensitivity experiments replace only $\phi_v$ while preserving the
rest of the EVLA architecture. The evaluated alternatives are TinyCNN,
ResNet-18, ConvNeXt-Tiny, Swin-Tiny, and a SimCLR-style ResNet-18. TinyCNN uses
five $3\times3$ convolutional layers with channels
$32\rightarrow64\rightarrow128\rightarrow128\rightarrow256$, stride 2,
padding 1, batch normalization, and ReLU. ResNet-18 projects a 512-dimensional
feature representation to 256; ConvNeXt-Tiny and Swin-Tiny project
768-dimensional representations to 256. Swin-Tiny uses patch size 4, window
size 7, and a $224\times224$ input. These alternatives are backbone ablations,
not changes to the multimodal fusion mechanism shown in
Fig.~\ref{fig:evla_overview}.

\subsection{Language Encoder and Loader Objective Remapping}
\label{subsec:language_encoder}

The reported main configuration uses the simple language encoder. The input
instruction is lowercased, tokenized by whitespace after stripping common
punctuation, and truncated or padded to 16 tokens. A vocabulary of size 1,000
is used; selected energy-related words have fixed entries and out-of-vocabulary
words are assigned deterministic indices by a character-sum rule. Each token is
mapped to a 64-dimensional learned embedding. A learned positional embedding is
added, the 16 token vectors are average-pooled, and the result is projected to
\begin{equation}
    e_i^l=\phi_l(L_i)\in\mathbb{R}^{256}.
    \label{eq:language_embedding}
\end{equation}

Before encoding, the loader infers an internal objective $O_i$ from the stored
instruction. The implemented internal objective set is
\begin{equation}
\begin{aligned}
\mathcal O_{\mathrm{internal}}=\{&\texttt{minimize\_cost},\texttt{minimize\_peak},
\texttt{maximize\_solar},\\
 &\texttt{grid\_support},\texttt{emergency\_backup},
\texttt{balance\_grid}\}.
\end{aligned}
\label{eq:internal_objectives}
\end{equation}
The keyword rules are evaluated in order: ``peak'' maps to
\texttt{minimize\_peak}; the joint presence of ``solar'' and ``self'' maps to
\texttt{maximize\_solar}; qualifying grid-support wording maps to
\texttt{grid\_support}; backup/emergency wording maps to
\texttt{emergency\_backup}; ``balance'' maps to \texttt{balance\_grid}; and
remaining instructions map to \texttt{minimize\_cost}. When language is enabled,
the loader selects a paraphrase associated with the inferred objective and may
append a short qualifier with probability 0.1. Thus, the evaluated language
pathway includes the objective-remapping stage shown in Fig.~\ref{fig:evla_inputs}
so the reported language experiments evaluate this remapped and paraphrased pathway rather than unseen-paraphrase generalization.

Optional language backbones are also implemented. The bidirectional-GRU option
uses 128-dimensional token embeddings, a 256-unit bidirectional GRU, and a
$512\rightarrow256$ projection. The BERT option uses
\texttt{bert-base-uncased}, maximum sequence length 16, and a
$768\rightarrow256$ projection; BERT remains frozen unless fine-tuning is
explicitly enabled. The main reported configuration uses the simple encoder.

\subsection{Vision--Language Action-Token Conditioning}
\label{subsec:action_conditioning}

\begin{figure}
    \centering
    \includegraphics[width=0.92\textwidth]{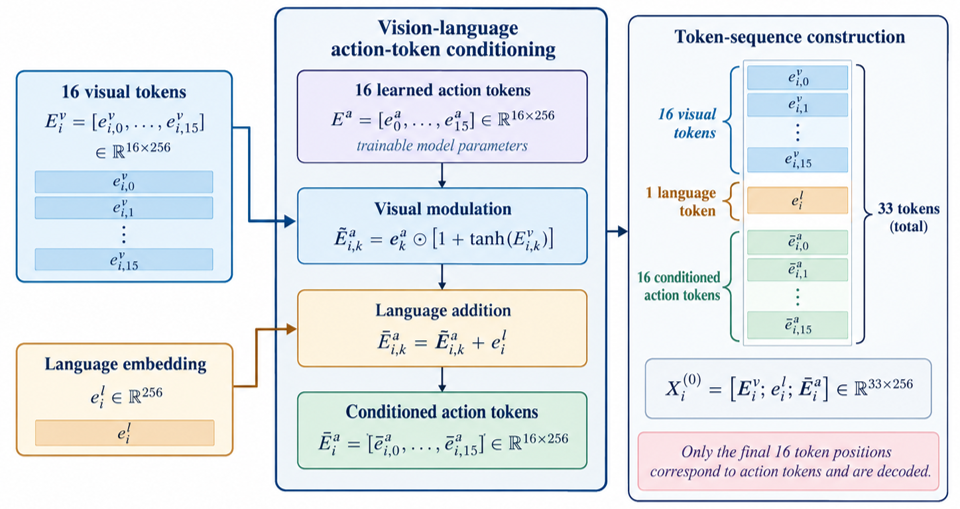}
    \caption{Vision--language conditioning of the learned action tokens and
    construction of the Transformer input sequence. Each action token is
    modulated by the visual token at the corresponding horizon position and
    then receives the same language embedding. The final sequence contains 16
    visual tokens, one language token, and 16 conditioned action tokens, for a
    total shape of $33\times256$. Only the final 16 positions are decoded.}
    \label{fig:evla_action_conditioning}
\end{figure}

The core action representation is shown in
Fig.~\ref{fig:evla_action_conditioning}. The model contains $H=16$ learned
horizon-specific action embeddings
\begin{equation}
    E^a
    =
    \left[e_0^a,\ldots,e_{15}^a\right]
    \in\mathbb{R}^{16\times256},
    \label{eq:action_tokens}
\end{equation}
which are trainable model parameters. These tokens act as horizon queries; the
trajectory is produced jointly rather than by iterative autoregressive rollout.

At horizon position $k$, the corresponding visual token modulates the learned
action embedding element-wise:
\begin{equation}
    \widetilde E^a_{i,k}
    =
    e_k^a\odot\left[1+\tanh(e_{i,k}^v)\right],
    \qquad k=0,\ldots,15.
    \label{eq:visual_action_conditioning}
\end{equation}
Since $1+\tanh(x)\in(0,2)$ for finite $x$, this stage provides bounded
multiplicative visual modulation. The language embedding is then added to every
horizon position,
\begin{equation}
    \overline E^a_{i,k}
    =
    \widetilde E^a_{i,k}+e_i^l,
    \label{eq:language_action_conditioning}
\end{equation}
which supplies common instruction-level intent while preserving the
step-specific visual modulation. The 16 conditioned action tokens are
$\overline E_i^a\in\mathbb{R}^{16\times256}$.

The Transformer input sequence is
\begin{equation}
    X_i^{(0)}
    =
    \left[E_i^v;\,e_i^l;\,\overline E_i^a\right]
    \in\mathbb{R}^{33\times256},
    \label{eq:token_sequence}
\end{equation}
where the first 16 positions are visual tokens, position 17 is the language
token, and the final 16 positions are conditioned action tokens. The mean visual
embedding is intentionally excluded from this sequence; it is used only in the
FiLM context and auxiliary visual pathway.

\subsection{FiLM Conditioning and Transformer Sequence Model}
\label{subsec:film_transformer}

\begin{figure}
    \centering
    \includegraphics[width=0.93\textwidth]{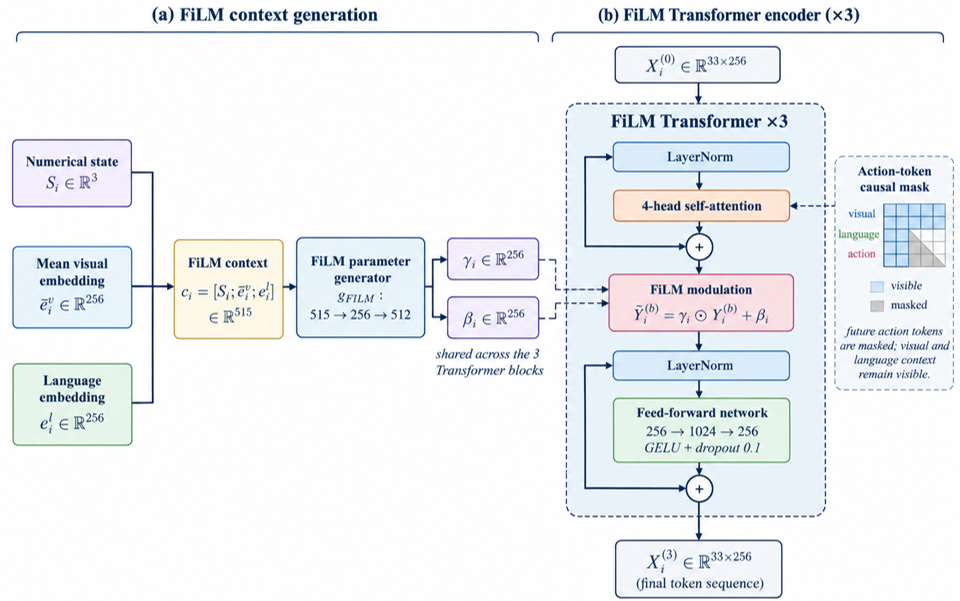}
    \caption{Multimodal FiLM conditioning and Transformer sequence model. The
    FiLM context concatenates the 3-D numerical state, the 256-D mean visual
    embedding, and the 256-D language embedding, giving 515 dimensions. An MLP
    $515\rightarrow256\rightarrow512$ produces 256-D scale and shift vectors.
    The $33\times256$ token sequence is processed by three identical
    Transformer blocks with four-head self-attention, an action-token causal
    mask, FiLM modulation, and an FFN of size $256\rightarrow1024\rightarrow256$.
    Future action-token positions are masked, whereas visual and language
    context remain visible.}
    \label{fig:evla_film_transformer}
\end{figure}

Figure~\ref{fig:evla_film_transformer} shows both the FiLM context generator and
one of the three repeated Transformer blocks. FiLM provides a second fusion path
that combines the explicit numerical state, global visual context, and language
context:
\begin{equation}
    c_i
    =
    \left[S_i;\overline e_i^v;e_i^l\right]
    \in\mathbb{R}^{3+256+256}
    =
    \mathbb{R}^{515}.
    \label{eq:film_context}
\end{equation}
A shared parameter generator produces scale and shift vectors,
\begin{equation}
\begin{aligned}
g_{\mathrm{FiLM}}(c_i)
&=[\gamma_i;\beta_i],\\
g_{\mathrm{FiLM}}
&:515\rightarrow256\rightarrow512,\\
\gamma_i,\beta_i
&\in\mathbb{R}^{256}.
\end{aligned}
\label{eq:film_generator}
\end{equation}

The same pair $(\gamma_i,\beta_i)$ is broadcast over token positions and shared
across all three Transformer blocks.

The sequence model uses $B=3$ Transformer-style blocks. Each block has four-head
self-attention, layer normalization, residual connections, and an FFN
$256\rightarrow1024\rightarrow256$ with GELU and dropout probability 0.1. For
block $b\in\{0,1,2\}$,
\begin{align}
    Y_i^{(b)}
    &=X_i^{(b)}+
    \operatorname{Drop}\!\left[
        \operatorname{MHA}\!\left(\operatorname{LN}(X_i^{(b)})\right)
    \right],
    \label{eq:attn_block}\\
    \widetilde Y_i^{(b)}
    &=\gamma_i\odot Y_i^{(b)}+\beta_i,
    \label{eq:film_block}\\
    X_i^{(b+1)}
    &=\widetilde Y_i^{(b)}+
    \operatorname{Drop}\!\left[
        \operatorname{FFN}\!\left(\operatorname{LN}(\widetilde Y_i^{(b)})\right)
    \right].
    \label{eq:ffn_block}
\end{align}
The causal mask is restricted to the action-token sub-sequence: action token
$k$ cannot attend to a future action token $h>k$, but visual and language
context tokens remain visible. Despite this causal structure, EVLA predicts all
16 actions in a single forward pass; it is not an iterative closed-loop or
step-by-step autoregressive controller.

During training, the complete visual embedding pathway and the complete
language embedding pathway are independently zeroed with probability 0.1. This
modality dropout regularizes the fusion mechanism and exposes the model to
missing-modality conditions during training.

\subsection{Residual Decoder and Deterministic Baseline}
\label{subsec:decoder_baseline}

\begin{figure}
    \centering
    \includegraphics[width=0.97\textwidth]{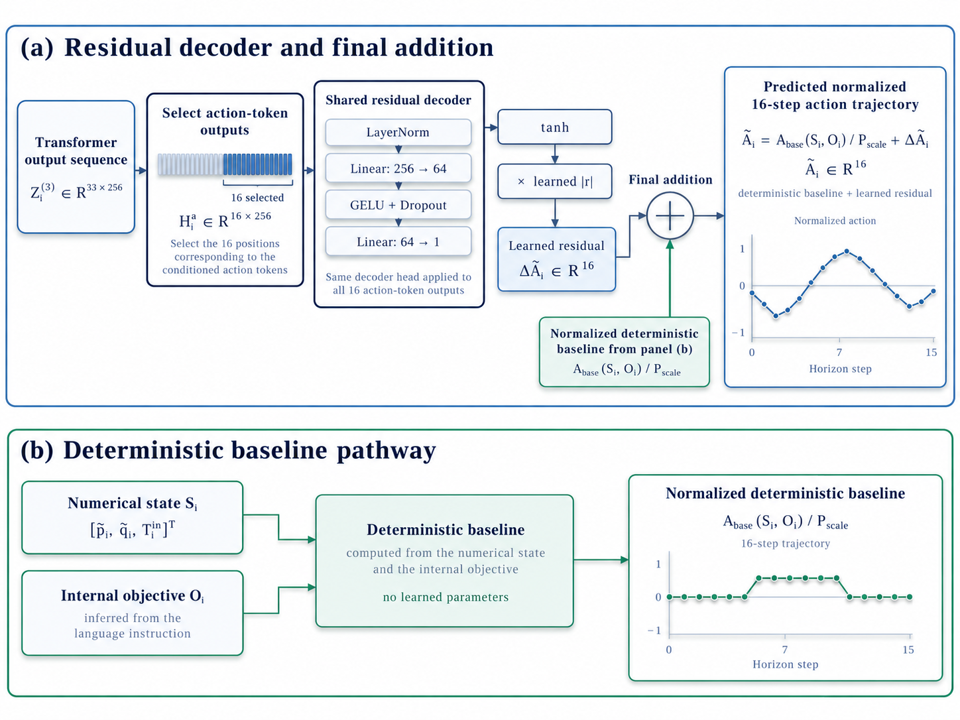}
    \caption{Action decoding and deterministic baseline. The final 16
    action-token representations are processed by one shared residual head and
    converted to a bounded learned residual. In parallel, the numerical state
    and loader-inferred internal objective determine a non-learned baseline
    action that is repeated over the 16-step horizon and normalized by
    $P_{\mathrm{scale}}$. The final normalized trajectory is obtained by adding
    the baseline and learned residual. The figure intentionally shows the
    baseline pathway structurally; the complete rule set is specified in
    Eq.~\eqref{eq:base_policy}.}
    \label{fig:evla_decoder_baseline}
\end{figure}

After the third Transformer block, only the 16 token positions corresponding to
conditioned action tokens are selected. Figure~\ref{fig:evla_decoder_baseline}
separates the learned residual route from the deterministic baseline route.

For each selected action-token representation $E_{i,k}^{\mathrm{out}}$, the
shared scalar head is
\begin{equation}
    \phi_{\mathrm{head}}:256\rightarrow64\rightarrow1,
    \label{eq:decoder_head}
\end{equation}
with layer normalization, GELU, and dropout. Its bounded residual output is
\begin{equation}
    \Delta\widetilde a_{i,k}
    =
    |r|\tanh\!\left(\phi_{\mathrm{head}}(E_{i,k}^{\mathrm{out}})\right),
    \qquad
    \Delta\widetilde A_i\in\mathbb{R}^{16},
    \label{eq:residual_output}
\end{equation}
where $r$ is a learned scalar initialized to 1. For fixed $r$,
$|\Delta\widetilde a_{i,k}|\leq |r|$.

The deterministic baseline receives the numerical state $S_i$ and the internal
objective $O_i$. It has no learned parameters. Let $\widetilde p_i$ and $q_i$
denote the preprocessed price and SOC. The normalized rule-based action is
\begin{equation}
b_i=
\begin{cases}
-0.8,&O_i=\texttt{minimize\_peak},\ \widetilde p_i>0.6,\ q_i>0.3,\\
\phantom{-}0.8,&O_i=\texttt{minimize\_peak},\ \widetilde p_i<0.2,\ q_i<0.9,\\
\phantom{-}1.0,&O_i=\texttt{maximize\_solar},\ \widetilde p_i<0.15,\ q_i<0.95,\\
-0.3,&O_i=\texttt{maximize\_solar},\ \widetilde p_i>0.5,\ q_i>0.1,\\
-0.9,&O_i=\texttt{grid\_support},\ \widetilde p_i>0.5,\ q_i>0.4,\\
\phantom{-}0.4,&O_i=\texttt{grid\_support},\ \widetilde p_i<0.25,\ q_i<0.8,\\
\phantom{-}0.3,&O_i=\texttt{emergency\_backup},\ \widetilde p_i<0.3,\ q_i<0.9,\\
-0.2,&O_i=\texttt{emergency\_backup},\ \widetilde p_i>0.6,\ q_i>0.5,\\
\phantom{-}0.5,&\mathrm{default},\ \widetilde p_i<0.3,\ q_i<0.8,\\
-0.5,&\mathrm{default},\ \widetilde p_i>0.7,\ q_i>0.2,\\
\phantom{-}0,&\mathrm{otherwise}.
\end{cases}
\label{eq:base_policy}
\end{equation}
The \texttt{balance\_grid} objective has no separate baseline rule and therefore
uses the default branch. The baseline trajectory is
\begin{equation}
    A_{\mathrm{base}}(S_i,O_i)
    =
    P_{\mathrm{scale}}[b_i,\ldots,b_i]
    \in\mathbb{R}^{16}.
    \label{eq:base_trajectory}
\end{equation}
The final model output is
\begin{equation}
    \widehat{\widetilde A}_i
    =
    \frac{A_{\mathrm{base}}(S_i,O_i)}{P_{\mathrm{scale}}}
    +
    \Delta\widetilde A_i.
    \label{eq:residual_prediction}
\end{equation}
No final clipping is applied after this addition. Consequently, the learned
prediction can exceed the nominal normalized interval, and the training loss
includes an explicit action-limit penalty. Architectural ablations additionally
include a no-residual mode, an MLP-fusion mode, and a base-only mode.

\subsection{Composite Training Objective}
\label{subsec:training_loss}

Losses are evaluated after mapping normalized predictions and targets back to
the stored-action scale,
\begin{equation}
    \hat a_{i,k}^{u}
    =P_{\mathrm{scale}}\hat{\widetilde a}_{i,k},
    \qquad
    a_{i,k}^{\star,u}
    =P_{\mathrm{scale}}\widetilde a_{i,k}^{\star}.
    \label{eq:loss_scale}
\end{equation}
The total objective is
\begin{equation}
\begin{aligned}
    \mathcal L_{\mathrm{total}}
    ={}&
    \mathcal L_{\mathrm{MSE}}
    +0.5\mathcal L_{\mathrm{smooth}}
    +0.1\mathcal L_{\mathrm{inactive}}\\
    &+2.0\mathcal L_{\mathrm{SOC}}
    +10.0\mathcal L_{\mathrm{limit}}
    +\mathcal L_{\mathrm{vision}}.
\end{aligned}
\label{eq:total_loss}
\end{equation}
The individual terms are as follows.
\begin{itemize}
    \item \textbf{Masked trajectory regression.}
    \begin{equation}
        \mathcal L_{\mathrm{MSE}}
        =
        \frac{\sum_{i,k}m_{i,k}
        (\hat a_{i,k}^{u}-a_{i,k}^{\star,u})^2}
        {\sum_{i,k}m_{i,k}}.
        \label{eq:loss_mse}
    \end{equation}

    \item \textbf{Smoothness.}
    \begin{equation}
        \mathcal L_{\mathrm{smooth}}
        =
        \frac{\sum_{i,k}m_{i,k}
        |\hat a_{i,k}^{u}-\hat a_{i,k-1}^{u}|}
        {\sum_{i,k}m_{i,k}},
        \qquad
        \hat a_{i,-1}^{u}=u_{i,0}.
        \label{eq:loss_smooth}
    \end{equation}
    If at least one objective label in the batch contains ``peak,'' the
    smoothness term for the entire mini-batch is multiplied by 1.5. Consequently,
    this weighting is batch-composition-dependent rather than applied only to the
    individual samples whose internal objective contains ``peak.''

    \item \textbf{Inaction penalty.} Near-zero predictions are penalized when
    the target magnitude exceeds $0.1P_{\mathrm{scale}}$. Because the reference trajectory generator
    candidate ranges depend on \(Z\), this threshold is an implementation-level
    regularizer and is not a regime-invariant physical criterion:
    \begin{equation}
        \mathcal L_{\mathrm{inactive}}
        =
        \frac{1}{\sum_{i,k}m_{i,k}}
        \sum_{i,k}m_{i,k}
        \exp\!\left(-\frac{|\hat a_{i,k}^{u}|}{0.5}\right)
        \mathbb I\!\left(|a_{i,k}^{\star,u}|>0.1P_{\mathrm{scale}}\right).
        \label{eq:loss_inactive}
    \end{equation}

    \item \textbf{SOC regularization.} The training-only SOC proxy is
    \begin{equation}
        \hat q_{i,k+1}
        =
        q_i+\sum_{j=0}^{k}
        \frac{\eta\hat a_{i,j}^{u}\Delta t}{C_{\mathrm{bat}}},
        \qquad
        \eta=0.95,\quad
        \Delta t=0.25,\quad
        C_{\mathrm{bat}}=13.5.
        \label{eq:soc_proxy}
    \end{equation}
    This regularizer uses the fixed value \(\eta=0.95\) for every training
    sample. It therefore does not reproduce the reference generator's regime-dependent
    efficiencies \(0.95\), \(0.80\), and \(0.60\); it should be interpreted as
    a generic training regularizer rather than a reference-generator-consistent feasibility
    model. The desired interval is $[0.1,0.9]$:
    \begin{equation}
        \mathcal L_{\mathrm{SOC}}
        =
        \frac{1}{\sum_{i,k}m_{i,k}}
        \sum_{i,k}m_{i,k}
        \left[
        \operatorname{ReLU}(0.1-\hat q_{i,k})
        +\operatorname{ReLU}(\hat q_{i,k}-0.9)
        \right]^2.
        \label{eq:loss_soc}
    \end{equation}

    \item \textbf{Action-limit regularization.} Stored-scale actions outside
    $[-5,5]$ are discouraged by
    \begin{equation}
        \mathcal L_{\mathrm{limit}}
        =
        \frac{\sum_{i,k}m_{i,k}
        \operatorname{ReLU}(|\hat a_{i,k}^{u}|-5)^2}
        {\sum_{i,k}m_{i,k}}.
        \label{eq:loss_limit}
    \end{equation}

    \item \textbf{Auxiliary visual objective.} When vision is enabled,
    \begin{equation}
    \begin{aligned}
        \mathcal L_{\mathrm{vision}}
        ={}&
        \operatorname{MSE}\!\left(
            \widehat F^{\mathrm{load}},F^{\mathrm{load}}/12000
        \right)\\
        &+
        \operatorname{MSE}\!\left(
            \widehat F^{\mathrm{solar}},F^{\mathrm{solar}}/8000
        \right).
    \end{aligned}
    \label{eq:loss_vision}
    \end{equation}
\end{itemize}
Optional contrastive and gate-regularization terms are implemented but have
zero weight in the reported configuration and therefore do not contribute to
Eq.~\eqref{eq:total_loss}.

\subsection{Reproducibility Summary of Model Dimensions}
\label{subsec:model_dimensions}

Table~\ref{tab:evla_model_dimensions} collects the fixed dimensions and
architecture constants used above so that the forward path can be reconstructed
without extracting values from multiple paragraphs.

\begin{table}
\centering
\caption{Key EVLA trajectory-prediction dimensions and fixed configuration
values used in the reported architecture.}
\label{tab:evla_model_dimensions}
\begin{tabular}{lll}
\toprule
Component & Quantity & Value \\
\midrule
Input image & native RGB size & $256\times128\times3$ \\
Visual preprocessing & network input & $224\times224\times3$ \\
Trajectory horizon & $H$ & 16 \\
Common latent width & $d$ & 256 \\
Visual representation & $E_i^v$ & $16\times256$ \\
Mean visual embedding & $\overline e_i^v$ & 256 \\
Simple language tokenizer & maximum length & 16 tokens \\
Simple language vocabulary & size & 1,000 \\
Simple language token embedding & width & 64 \\
Language representation & $e_i^l$ & 256 \\
Numerical state & $S_i$ & 3 \\
Learned action-token bank & $E^a$ & $16\times256$ \\
Transformer input & $X_i^{(0)}$ & $33\times256$ \\
Transformer depth & blocks & 3 \\
Self-attention & heads & 4 \\
Feed-forward network & hidden width & $256\rightarrow1024\rightarrow256$ \\
Transformer dropout & probability & 0.1 \\
Modality dropout & vision/language & 0.1 independently \\
FiLM context & $c_i$ & 515 \\
FiLM generator & MLP & $515\rightarrow256\rightarrow512$ \\
FiLM outputs & $\gamma_i,\beta_i$ & $256+256$ \\
Residual decoder & shared MLP & $256\rightarrow64\rightarrow1$ \\
Residual scale & learned $r$ & initialized to 1 \\
Action scale & $P_{\mathrm{scale}}$ & 5 \\
SOC regularizer & $(\eta,\Delta t,C_{\mathrm{bat}})$ & $(0.95,0.25,13.5)$ \\
Desired SOC interval & & $[0.1,0.9]$ \\
Nominal stored-action interval & & $[-5,5]$ \\
\bottomrule
\end{tabular}
\end{table}

\subsection{Inference Behavior and Modality Routing}
\label{subsec:model_interpretation}

At inference time, EVLA receives \((V_i,S_i,L_i)\) and predicts all 16 normalized actions in one forward pass. The action-token causal mask restricts information exchange among future action positions inside the Transformer, but the model is not rolled forward autoregressively and does not update the residential state between predicted steps. The SOC recurrence in Eq.~\eqref{eq:soc_proxy} enters only through the training objective.

Three mechanisms bound or regularize the learned trajectory: target normalization in Eq.~\eqref{eq:target_normalization}, multiplicative visual modulation in Eq.~\eqref{eq:visual_action_conditioning}, and the learned residual scale in Eq.~\eqref{eq:residual_output}. Together with the SOC and action-limit penalties, they constrain the numerical learning problem without changing the one-shot inference structure.

The ablations follow the implemented routing directly. Removing vision suppresses the complete energy-field representation, including the regime texture; removing language suppresses the processed text embedding; and removing state suppresses the three-dimensional numerical input used by FiLM and the deterministic baseline. The resulting comparisons quantify dependence on these implemented inputs, while paired physical or semantic interventions require the additional controls discussed in Sections~\ref{sec:discussion} and~\ref{sec:limitations}.
\section{Experimental Protocol}
\label{sec:experimental_protocol}

This section describes the protocol used to evaluate EVLA as a multimodal
trajectory-prediction benchmark. The experiments examine the effects of visual
input, processed language, numerical state, visual-backbone selection, fusion
design, and residual prediction on agreement with the generated reference
trajectories. The protocol focuses on agreement with the generated reference trajectories, modality sensitivity, computational cost, and variation across training seeds.

\subsection{Computing Environment}

All experiments were executed on a workstation running Ubuntu 22.04.5 LTS,
with an Intel Core Ultra 9 275HX processor, 62~GiB of system memory, and an
NVIDIA GeForce RTX 5090 GPU with 24~GB of memory. The system used NVIDIA
driver 580.159.03, Python 3.10.12, and PyTorch 2.9.1 with CUDA 12.8 support;
the driver reported CUDA 13.0. Versions of the remaining dependencies are
provided in the supplementary material.

All neural configurations were evaluated on the same workstation. The timing
values are therefore useful for comparison within this study but should not be
interpreted as platform-independent latency measurements.

\subsection{Dataset Partitioning and Experimental Subset}
\label{subsec:data_split}

The complete EVLA collection contains \(439{,}203\) base operating windows and
\(6{,}588{,}045\) generated instances. The model study uses a fixed subset of
\(N_{\mathrm{train}}=5{,}000\), \(N_{\mathrm{val}}=500\), and
\(N_{\mathrm{test}}=500\) instances.

The available JSON files are grouped according to their household directory.
Within each household, filenames are sorted lexicographically and divided
using proportions \(r_{\mathrm{train}}=0.70\),
\(r_{\mathrm{val}}=0.15\), and \(r_{\mathrm{test}}=0.15\).
Lexicographic ordering was intended to preserve source order, but the
implementation does not parse and verify physical timestamps. The procedure
is therefore described as a household-grouped ordered split rather than a
formally verified chronological split.

The three candidate partitions are shuffled independently using seed 42.
The first 5,000 training, 500 validation, and 500 test files are then retained.
The subset-selection seed remains fixed across configurations.

Partitioning is performed at the generated-file level rather than by the
base-state \texttt{group\_id}. The implementation therefore does not guarantee
that all 15 variants of a base operating window remain in the same partition.
No purge interval is applied between potentially overlapping source windows.
Accordingly, the current split is not claimed to be base-group-disjoint,
overlap-free, or leakage-free.

The selected 6,000 instances represent approximately \(0.091\%\) of the
complete generated collection. This fraction is intentionally a pilot
experimental sample for the multi-configuration study; the results consequently
characterize that controlled subset and do not establish performance after
training on the complete benchmark.

\subsection{Experimental Configurations}

The evaluation contains 36 named configurations. Some labels repeat the same
effective architecture under different experimental groupings, so 36 denotes
the number of evaluated configuration labels rather than the number of
completely distinct architectures.

\subsubsection{Modality ablations}

For each of six visual backbones, four modality settings are evaluated:

\begin{enumerate}[leftmargin=*]
    \item full input: vision, processed language, and numerical state;
    \item no vision: processed language and numerical state;
    \item no language: vision and numerical state; and
    \item state only: numerical state without vision or language.
\end{enumerate}

The six backbones and four modality settings produce 24 named modality
configurations. A no-vision result removes the complete energy-field image; it
does not isolate the hidden texture from the remaining image information.
Likewise, the language settings use the objective remapping described in
Section~\ref{sec:controller} and therefore evaluate the executed
processed-language pathway rather than an unmodified five-class protocol.

\subsubsection{Visual-backbone comparison}

Six full-modality entries compare TinyCNN, MobileNetV3-Small, ResNet-18,
ConvNeXt-Tiny, Swin-Tiny, and the SimCLR-style encoder. They use the same
downstream embedding dimension and trajectory horizon. These entries overlap
with the corresponding full models in the modality-ablation group.

\subsubsection{Architecture variants}

Four MobileNet-based entries evaluate the full attention and residual
architecture, prediction without the deterministic residual baseline, direct
MLP fusion, and the deterministic base-only policy. The full MobileNet entry
also overlaps with equivalent full-modality entries in the other groups.

\subsubsection{Neural baselines}

An MLP and an LSTM \citep{hochreiter1997lstm} are evaluated without the RGB
image or language encoder, but they are not input-identical baselines. The MLP
maps only the three-dimensional numerical state directly to the 16-step action
trajectory. The LSTM initializes its recurrent state from the same
three-dimensional numerical state and, at each prediction step, receives a
five-dimensional input formed by the numerical state, its own previous predicted
action (initialized to zero at the first step), and the normalized step index
\(t/H\). Thus, the LSTM has explicit recurrent action-history and horizon-position
signals that are not supplied as separate numerical inputs to the main EVLA
predictor. These models should therefore be interpreted as numerical/state-derived
reference architectures rather than strictly input-matched controls.

The 36 labels comprise 24 modality configurations, six backbone entries, four
architecture entries, and two neural baselines.

\subsection{Training Procedure}

Each learned configuration is trained on the 5,000-instance training subset.
Validation MSE is monitored for early stopping, while the base-only
configuration requires no training.

The optimizer is AdamW \citep{loshchilov2019adamw} with an initial learning
rate of \(3\times10^{-4}\), weight decay \(0.01\), and mini-batch size 32.
Training continues for at most 100 epochs. Gradients are clipped to a norm of
1.0, and both network dropout and modality dropout use probability 0.1. The
cosine-annealing warm-restart scheduler \citep{loshchilov2017sgdr} uses
\(T_0=10\) and \(T_{\mathrm{mult}}=2\) and is advanced once per epoch. Training stops after
10 epochs without improvement in validation MSE.

Training uses the composite objective defined in
Eq.~\eqref{eq:total_loss}, rather than MSE alone. Its weights are
\(w_{\mathrm{MSE}}=1.0\), \(w_{\mathrm{smooth}}=0.5\),
\(w_{\mathrm{inactive}}=0.1\), \(w_{\mathrm{SOC}}=2.0\),
\(w_{\mathrm{limit}}=10.0\), and \(w_{\mathrm{vision}}=1.0\).
Optional contrastive and gate-regularization terms have zero weight.
Automatic mixed precision is not used.

A checkpoint is saved whenever validation MSE improves. In the executed runs,
the training function does not reload the lowest-validation-MSE checkpoint before
testing; the reported test result is produced by the in-memory model at the
stopping epoch. The reported numbers therefore characterize the executed
early-stopping implementation rather than a best-checkpoint-selected protocol.
A corrected future evaluation should reload the validation-selected checkpoint
before the final test pass.

The experiment uses seeds 42, 43, and 44. The same data subset is used for all
runs. Evaluating 36 named configurations with three seeds gives 108
configuration--seed evaluations.

\subsection{Diagnostic Input Interventions}

Three input interventions are implemented for the learned multimodal
configurations. In the language-scrambling diagnostic, processed instructions
are permuted within each test mini-batch while the numerical state, image, and
internal objective label remain unchanged. This intervention changes the text
but not every objective-associated model input.

In the cross-batch visual-scrambling diagnostic, images from the current batch
are combined with images retained from a preceding batch and randomly
reassigned. The first test batch is not modified. In the zero-vision
diagnostic, the RGB image is replaced by zero while the remaining inputs are
preserved.

The saved intervention gap is
\(G=E_{\mathrm{nominal}}-E_{\mathrm{intervention}}\). A negative value
therefore means that the intervention increased MSE, whereas a positive value
means that it reduced MSE.

Although the configuration includes disturbance-related fields, the supplied
loader does not apply a disturbance operator during sample loading.
Disturbance-labelled results are therefore excluded from the reported
evidence.

\subsection{Trajectory-Error Metrics}

The evaluator reconstructs the stored-action scale using
\(\hat{a}^{\,u}_{i,k}=P_{\mathrm{scale}}\hat{a}_{i,k}\) and
\(a^{\star,u}_{i,k}=P_{\mathrm{scale}}a^\star_{i,k}\), where
\(P_{\mathrm{scale}}=5\). Let \(m_{i,k}\) be the target-validity mask. The
trajectory metrics are

\begin{align}
    \mathrm{MSE}
    &=
    \frac{
        \sum_{i,k}m_{i,k}
        \left(
            \hat{a}^{\,u}_{i,k}-a^{\star,u}_{i,k}
        \right)^2
    }{
        \sum_{i,k}m_{i,k}
    },
    \nonumber\\
    \mathrm{MAE}
    &=
    \frac{
        \sum_{i,k}m_{i,k}
        \left|
            \hat{a}^{\,u}_{i,k}-a^{\star,u}_{i,k}
        \right|
    }{
        \sum_{i,k}m_{i,k}
    },
    \qquad
    \mathrm{RMSE}=\sqrt{\mathrm{MSE}}.
    \label{eq:trajectory_metrics}
\end{align}

MSE is expressed in squared stored-action units, while MAE and RMSE are in
stored-action units. These metrics should not be reported in
\(\mathrm{kW}^2\) or \(\mathrm{kW}\) unless the physical action unit and sign
convention are independently established.

\subsection{Operational Diagnostic Metrics}
\label{subsec:operational_metrics}

The evaluator also calculates peak action, action throughput, an energy-cost
proxy, SOC-bound excess, and a self-consumption proxy. These are
implementation-defined diagnostics rather than validated physical or economic
quantities.

Let \(P_i^{\mathrm{load}}\) and \(P_i^{\mathrm{solar}}\) denote the scalar load
and solar-proxy values associated with instance \(i\). Peak action, the cycle
proxy, assumed grid import, and the energy-cost proxy are

\begin{align}
    P^{\mathrm{peak}}
    &=
    \frac{1}{N}
    \sum_i
    \max_k
    \left|
        \hat{a}^{\,u}_{i,k}
    \right|,
    \nonumber\\
    C_{\mathrm{cycle}}
    &=
    \frac{\Delta t}{N}
    \sum_i\sum_k
    \left|
        \hat{a}^{\,u}_{i,k}
    \right|,
    \nonumber\\
    P_{i,k}^{\mathrm{grid}}
    &=
    \max
    \left(
        \frac{P_i^{\mathrm{load}}-P_i^{\mathrm{solar}}}{1000}
        +
        \hat{a}^{\,u}_{i,k},
        0
    \right),
    \nonumber\\
    C_{\mathrm{energy}}
    &=
    \frac{\Delta t}{N}
    \sum_i p_i
    \sum_k P_{i,k}^{\mathrm{grid}},
    \qquad
    \Delta t=0.25.
    \label{eq:operational_proxies}
\end{align}

Despite its implementation name, \(C_{\mathrm{cycle}}\) is an
action-throughput proxy; it does not model degradation or count physical
battery cycles. Likewise, \(C_{\mathrm{energy}}\) is not an estimate of an
actual household electricity bill.

The SOC trajectory is reconstructed using
\(\hat{q}_{i,k+1}=q_i+\sum_{j=0}^{k}
\eta\hat{a}^{\,u}_{i,j}\Delta t/C_{\mathrm{bat}}\), with
\(\eta=0.95\), \(\Delta t=0.25\), and \(C_{\mathrm{bat}}=13.5\).
The reported SOC-bound diagnostic is

\begin{equation}
    V_{\mathrm{SOC}}
    =
    \frac{1}{NH}
    \sum_{i=1}^{N}
    \sum_{k=1}^{H}
    \left[
        \operatorname{ReLU}(0.1-\hat{q}_{i,k})
        +
        \operatorname{ReLU}(\hat{q}_{i,k}-0.9)
    \right].
    \label{eq:violation_metric}
\end{equation}

This quantity is the mean magnitude of SOC-bound excess, not a binary
violation rate. It does not include separate action or temperature violations.

The self-consumption proxy is implemented as

\begin{equation}
    R_{\mathrm{self}}
    =
    \frac{1}{N}
    \sum_i
    \operatorname{clip}
    \left(
        \frac{
            1000\Delta t
            \sum_k\max(-\hat{a}^{\,u}_{i,k},0)
        }{
            P_i^{\mathrm{load}}+10^{-6}
        },
        0,1
    \right).
    \label{eq:self_consumption_metric}
\end{equation}

It measures normalized negative-action throughput relative to a scalar load
value. It is not the conventional fraction of locally generated renewable
energy consumed on site.

\subsection{Model Size and Inference Timing}

Parameter count is the sum of trainable parameters; frozen pretrained
parameters are excluded. The tables report this quantity in millions.

Inference time is measured for each test mini-batch using
\texttt{time.perf\_counter}. The elapsed time is divided by the batch size, and
the mean of the resulting values is reported in milliseconds per instance.
The evaluation batch size is 32.

No explicit warm-up iterations or GPU synchronization calls are used.
Because CUDA operations can execute asynchronously, the timing values are
approximate pipeline diagnostics rather than synchronized GPU-latency
measurements.

\subsection{Statistical Reporting}

Each named configuration is evaluated with seeds 42, 43, and 44. For a metric
\(x\), the reported value is the arithmetic mean
\(\bar{x}=R^{-1}\sum_{r=1}^{R}x_r\) and sample standard deviation
\(s=[(R-1)^{-1}\sum_{r=1}^{R}(x_r-\bar{x})^2]^{1/2}\), with \(R=3\).

Three seeds provide only a limited estimate of training variability. No formal
hypothesis tests, confidence intervals, or corrections for multiple
comparisons are reported. Conclusions therefore emphasize the magnitude and
consistency of changes rather than fine-grained ranking. This is particularly
important for vision, because the aggregate difference between the full and
no-vision configurations is small, whereas language removal produces a much
larger change in trajectory error.
\section{Results}
\label{sec:results}

This section reports the dataset audit and the controlled ablation study. All learned-model results use the fixed subset of 5,000 training, 500 validation, and 500 test instances described in Section~\ref{subsec:data_split}. Prediction errors are reported in the evaluator's stored-action scale; the operational quantities are implementation-defined diagnostics as specified in Section~\ref{subsec:operational_metrics}.
\subsection{Dataset Audit}

The structural audit scanned all \(6{,}588{,}045\) generated JSON instances
and identified \(439{,}203\) unique base-state group identifiers. The resulting
expansion factor is exactly 15, consistent with the Cartesian construction
over three hidden regimes and five dataset-level language objectives.

The audit found zero malformed JSON records, zero records lacking a state
entry, and zero stored trajectories with lengths other than 16. When instances
were grouped by base-state identifier, no group contained inconsistent
battery-SOC or indoor-temperature metadata.

Image availability was checked for a sample of 6,588 generated instances, with
no missing image found in that sample. This is a sampled availability check,
not an exhaustive scan of all \(6{,}588{,}045\) image files.

The audit therefore supports the intended record structure and paired grouping of the generated metadata. Physical fidelity, language semantics, and partition leakage are separate evaluation questions and are addressed in Section~\ref{sec:limitations}.

\subsection{Visual-Backbone Comparison}

Table~\ref{tab:backbone} and
Fig.~\ref{fig:principal_ablations}(a) compare the six designated
full-modality backbone entries. MobileNet obtains the lowest mean MSE,
\(0.3829\pm0.0003\), followed by the SimCLR-style encoder at
\(0.3834\pm0.0005\), ResNet at \(0.3835\pm0.0004\), TinyCNN at
\(0.3846\pm0.0013\), ConvNeXt at \(0.3849\pm0.0019\), and Swin at
\(0.3854\pm0.0024\).

The difference between the lowest and highest mean MSE is only 0.0025.
Increasing visual-backbone capacity therefore does not produce a meaningful
aggregate-error improvement under the current subset and training protocol.

MobileNet contains approximately 4.44 million trainable parameters and has an
approximate timing diagnostic of \(0.495\)~ms per instance. TinyCNN contains
approximately 4.17 million trainable parameters and has a lower reported
timing of \(0.298\)~ms, with an MSE increase of approximately 0.0017.
ConvNeXt and Swin contain more than 19 million trainable parameters and have
reported timings above \(1.2\)~ms without reducing MSE.

MobileNet therefore has the lowest mean error in the designated backbone
comparison, whereas TinyCNN has the smallest reported timing among the
vision-enabled entries. The timing results remain approximate because the
evaluation does not use warm-up runs or explicit CUDA synchronization.

\begin{table}
\centering
\caption{Visual-backbone comparison aggregated over seeds 42, 43, and 44. Timing values are approximate per-instance diagnostics.}
\label{tab:backbone}
\small
\begin{tabular}{lcccc}
\toprule
Backbone & MSE & MAE & Params (M) & Time (ms) \\
\midrule
MobileNet & \(0.3829\pm0.0003\) & 0.485 & 4.44 & 0.495 \\
SimCLR    & \(0.3834\pm0.0005\) & 0.486 & 14.93 & 0.788 \\
ResNet    & \(0.3835\pm0.0004\) & 0.486 & 14.26 & 0.823 \\
TinyCNN   & \(0.3846\pm0.0013\) & 0.487 & 4.17 & 0.298 \\
ConvNeXt  & \(0.3849\pm0.0019\) & 0.487 & 19.36 & 1.282 \\
Swin      & \(0.3854\pm0.0024\) & 0.488 & 19.26 & 1.236 \\
\bottomrule
\end{tabular}
\end{table}

\subsection{MobileNet Modality Ablation}

Table~\ref{tab:modality_mobilenet} and
Fig.~\ref{fig:principal_ablations}(b) report the designated MobileNet modality
ablation. The full model produces an MSE of \(0.3856\pm0.0039\), whereas the
no-vision configuration produces \(0.3843\pm0.0013\). The no-vision mean is
0.0013 lower, and this difference is small relative to the full model's
seed-to-seed standard deviation. The experiment therefore provides no
aggregate-error evidence that including the RGB image improves trajectory
prediction.

Removing language has a substantially larger effect. The no-language
configuration produces an MSE of \(0.8628\pm0.0001\), while the state-only
configuration produces \(0.8626\pm0.0000\). Their MAE values are approximately
0.721 and 0.719, respectively, compared with approximately 0.488 and 0.487 for
the full and no-vision configurations.

The large error increase shows that the trained predictor depends strongly on the processed text pathway. Because the loader also reconstructs an internal objective used by the deterministic baseline, this ablation isolates the text-encoder contribution rather than a complete intervention on the original five-class objective variable.

The energy-cost proxy is approximately 0.042 for the full model, 0.040 for the
no-vision model, and 0.088 for the no-language and state-only models. The
SOC-bound excess is zero at the displayed precision for the full and no-vision
models and approximately 0.0022--0.0023 for the two configurations without
processed language.

By contrast, the image ablation leaves aggregate error essentially unchanged. A direct test of regime use therefore requires paired, group-aware comparisons in which the base state and processed language are fixed while only the hidden regime changes.

\begin{table}
\centering
\caption{MobileNet modality ablation aggregated over seeds 42, 43, and 44. Energy and violation values are internal benchmark diagnostics.}
\label{tab:modality_mobilenet}
\small
\begin{tabular}{lcccc}
\toprule
Variant & MSE & MAE & Energy proxy & Viol. \\
\midrule
Full        & \(0.3856\pm0.0039\) & 0.488 & 0.042 & 0.0000 \\
No vision   & \(0.3843\pm0.0013\) & 0.487 & 0.040 & 0.0000 \\
No language & \(0.8628\pm0.0001\) & 0.721 & 0.088 & 0.0022 \\
State only  & \(0.8626\pm0.0000\) & 0.719 & 0.088 & 0.0023 \\
\bottomrule
\end{tabular}
\end{table}

\subsection{Architecture Ablation}

\begin{figure}
    \centering

    \begin{minipage}[t]{0.315\textwidth}
        \centering
        \includegraphics[width=\linewidth]
        {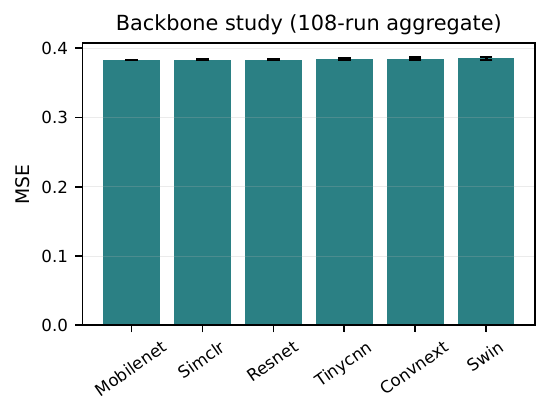}

        \vspace{1mm}
        {\small\textbf{(a) Visual-backbone comparison}}
    \end{minipage}
    \hfill
    \begin{minipage}[t]{0.315\textwidth}
        \centering
        \includegraphics[width=\linewidth]
        {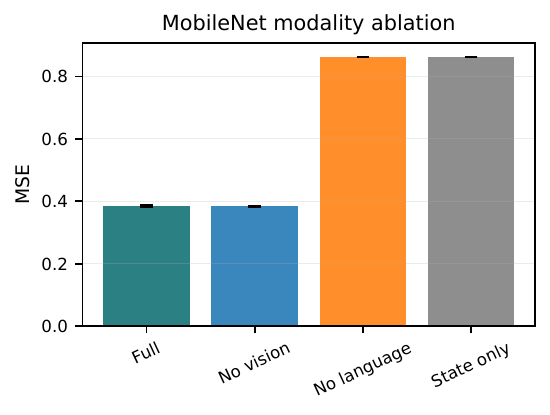}

        \vspace{1mm}
        {\small\textbf{(b) Modality ablation}}
    \end{minipage}
    \hfill
    \begin{minipage}[t]{0.315\textwidth}
        \centering
        \includegraphics[width=\linewidth]
        {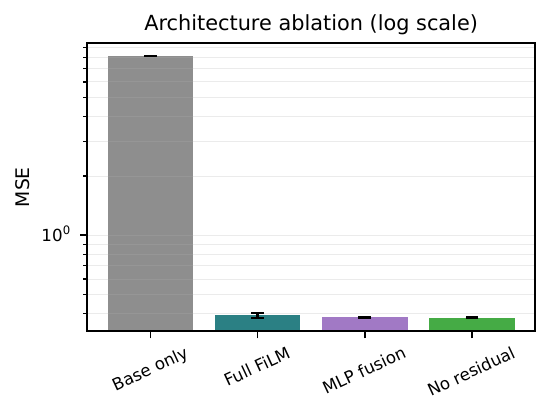}

        \vspace{1mm}
        {\small\textbf{(c) Architecture ablation}}
    \end{minipage}

    \caption{Principal EVLA ablations aggregated over seeds 42, 43, and 44:
    (a) trajectory MSE for the six designated full-modality visual backbones;
    (b) MobileNet-family modality ablation, showing strong dependence of the
    implemented multimodal pathway on processed language but no aggregate
    degradation after removing vision; and (c) MobileNet architecture ablation
    on a logarithmic scale, showing the substantially larger error of the
    deterministic base-only model. The language result is architecture-specific
    and should be interpreted together with the numerical-only baselines in
    Table~\ref{tab:baseline_comparison}.}
    \label{fig:principal_ablations}
\end{figure}

Table~\ref{tab:architecture_ablation} and
Fig.~\ref{fig:principal_ablations}(c) compare the MobileNet-based architecture
variants. The deterministic base-only model produces an MSE of 8.0795,
substantially exceeding the errors of the learned models. Its SOC-bound excess
is approximately 0.1084 in the complete result bundle. The deterministic
baseline is therefore not an adequate stand-alone approximation of the
generated targets.

Among the learned variants, the no-residual model produces an MSE of
\(0.3829\pm0.0001\), MLP fusion produces \(0.3832\pm0.0010\), and the full
residual FiLM model produces \(0.3922\pm0.0111\). Under the current protocol,
neither residual addition nor the complete attention--FiLM architecture
improves aggregate MSE relative to the simpler alternatives. The full model
also exhibits greater seed-to-seed variation.

The language-scrambling diagnostic provides a different perspective. Its
stored gap is
\(\Delta_{\mathrm{lang}}=\mathrm{MSE}_{\mathrm{nominal}}
-\mathrm{MSE}_{\mathrm{scrambled}}\). The full FiLM and MLP-fusion models have
mean gaps of approximately \(-0.927\) and \(-0.935\), respectively. Thus,
scrambling the text increases their MSE substantially. The no-residual model
has a gap of only approximately \(-0.008\), indicating considerably weaker
sensitivity to the text input.

Only the text strings are scrambled in this diagnostic; the internal objective
labels supplied to the model and deterministic baseline remain unchanged. The
test therefore measures sensitivity to the text encoder conditional on a
fixed internal objective rather than responsiveness to a complete
counterfactual change in intent.

The comparison nevertheless shows that low nominal MSE does not necessarily
imply strong dependence on the text embedding. The no-residual model obtains
the lowest mean error in this comparison while exhibiting the weakest
language-scrambling effect.

\begin{table}
\centering
\caption{Architecture ablation for MobileNet-based EVLA variants, aggregated over seeds 42, 43, and 44.}
\label{tab:architecture_ablation}
\small
\begin{tabular}{lcccc}
\toprule
Variant & MSE & Params (M) & Time (ms) & Lang. gap \\
\midrule
Base only   & \(8.0795\pm0.0000\) & 1.26 & 0.114 & -- \\
Full FiLM   & \(0.3922\pm0.0111\) & 4.44 & 0.490 & -0.927 \\
MLP fusion  & \(0.3832\pm0.0010\) & 1.46 & 0.453 & -0.935 \\
No residual & \(0.3829\pm0.0001\) & 4.44 & 0.490 & -0.008 \\
\bottomrule
\end{tabular}
\end{table}

\subsection{Neural Baseline Comparison}

Table~\ref{tab:baseline_comparison} compares the designated MobileNet backbone
entry with the MLP and LSTM baselines. MobileNet obtains an MSE of
\(0.3829\pm0.0003\), compared with \(0.3887\pm0.0001\) for the MLP and
\(0.3893\pm0.0004\) for the LSTM. The corresponding absolute differences are
approximately 0.0058 and 0.0064 and should be regarded as modest. In particular,
the strong degradation of the MobileNet-family no-language ablation should not
be interpreted as evidence that language is universally required to obtain low
aggregate MSE on this subset: numerical-only MLP and LSTM baselines remain close
to the designated MobileNet result. The ablation instead demonstrates dependence
of the implemented multimodal pathway on its processed-language route.

The MLP is substantially smaller, with approximately 0.07 million trainable
parameters, and has a reported timing diagnostic of \(0.008\)~ms per instance.
The LSTM contains approximately 1.06 million parameters and reports
\(0.134\)~ms, while MobileNet contains approximately 4.44 million parameters
and reports \(0.495\)~ms. These accuracy comparisons are not strictly
input-matched: the MLP uses only the three-dimensional numerical state, whereas
the LSTM additionally receives its own previous predicted action and an explicit
normalized time-step signal at each recurrent step. The LSTM result should
therefore not be read as a pure three-state-variable control with the same input
structure as the main EVLA model.

The energy-cost proxies are approximately 0.0347 for MobileNet, 0.0402 for the
MLP, and 0.0423 for the LSTM. No SOC-bound excess is reported at the displayed
precision for these entries. These operational quantities remain dependent on
the evaluator's assumed action scaling and sign convention.

\begin{table}
    \centering
    \caption{Comparison of the designated MobileNet entry with the numerical
    neural baselines, aggregated over seeds 42, 43, and 44. Timing values are
    approximate, and energy values are internal benchmark proxies.}
    \label{tab:baseline_comparison}
    \small
    \resizebox{\columnwidth}{!}{%
    \begin{tabular}{lccccc}
        \toprule
        Model & MSE & MAE & Params (M) & Time (ms) & Energy proxy \\
        \midrule
        MobileNet & \(0.3829\pm0.0003\) & 0.4850 & 4.44 & 0.495 & 0.0347 \\
        MLP       & \(0.3887\pm0.0001\) & 0.4905 & 0.07 & 0.008 & 0.0402 \\
        LSTM      & \(0.3893\pm0.0004\) & 0.4913 & 1.06 & 0.134 & 0.0423 \\
        \bottomrule
    \end{tabular}%
    }
\end{table}

Disturbance-labelled runs are omitted from the reported robustness evidence because the supplied loader does not apply the corresponding perturbation operator during sample loading.

\subsection{Lowest-MSE Named Configurations}

Table~\ref{tab:top10_configs} lists the ten named configurations with the lowest
mean MSE in the aggregated result bundle. The designated MobileNet backbone
entry and no-residual MobileNet entry both have a rounded mean MSE of 0.3829,
followed by MLP fusion at \(0.3832\pm0.0010\).

\begin{table}
\centering
\caption{Ten named configurations with the lowest mean MSE in the aggregated ablation study.}
\label{tab:top10_configs}
\small
\resizebox{\textwidth}{!}{%
\begin{tabular}{lrrrrrr}
\toprule
Variant & MSE & MAE & RMSE & Params (M) & Time (ms) & Lang. gap \\
\midrule
Backbone-Mobilenet & 0.3829$\pm$0.0003 & 0.485 & 0.619 & 4.44 & 0.495 & -0.893 \\
Arch-NoResidual-MobileNet & 0.3829$\pm$0.0001 & 0.485 & 0.619 & 4.44 & 0.490 & -0.008 \\
Arch-MLPFusion-MobileNet & 0.3832$\pm$0.0010 & 0.486 & 0.619 & 1.46 & 0.453 & -0.935 \\
Modality-NoVision-Resnet & 0.3833$\pm$0.0006 & 0.486 & 0.619 & 3.33 & 0.186 & -0.939 \\
Modality-Full-Convnext & 0.3834$\pm$0.0010 & 0.486 & 0.619 & 19.36 & 1.289 & -0.922 \\
Backbone-Simclr & 0.3834$\pm$0.0005 & 0.486 & 0.619 & 14.93 & 0.788 & -0.929 \\
Backbone-Resnet & 0.3835$\pm$0.0004 & 0.486 & 0.619 & 14.26 & 0.823 & -0.941 \\
Modality-NoVision-Swin & 0.3836$\pm$0.0003 & 0.486 & 0.619 & 3.33 & 0.187 & -0.928 \\
Modality-NoVision-Simclr & 0.3837$\pm$0.0003 & 0.486 & 0.619 & 3.33 & 0.189 & -0.936 \\
Modality-Full-Swin & 0.3840$\pm$0.0004 & 0.486 & 0.620 & 19.26 & 1.244 & -0.943 \\
\bottomrule
\end{tabular}%
}
\end{table}

Several no-vision entries also appear among the ten lowest-MSE configurations:
no-vision ResNet at \(0.3833\pm0.0006\), no-vision Swin at
\(0.3836\pm0.0003\), and no-vision SimCLR at \(0.3837\pm0.0003\). Their
presence is consistent with the modality ablation, which does not show an
aggregate-error benefit from vision.

The table includes separately named entries with equivalent effective
settings. For example, \texttt{Backbone-Mobilenet},
\texttt{Modality-Full-Mobilenet}, and \texttt{Arch-Full-MobileNet} describe
the same principal full-modality MobileNet architecture but were trained at
different positions in the experiment loop. Their mean MSE values are
approximately 0.3829, 0.3856, and 0.3922, respectively.

Because the random-number generators are not reset immediately before every
named configuration, equivalent entries do not necessarily receive identical
parameter initializations or mini-batch sequences. Their variation indicates
that exact ranking among models with closely spaced MSE values should not be
overinterpreted.

Table~\ref{tab:top10_configs} should therefore be understood as a list of
lowest-MSE named entries rather than a definitive ranking of distinct
architectures.

\section{Discussion}
\label{sec:discussion}

The experiments reveal a clear asymmetry in how the current EVLA predictor uses its inputs. Processed language carries a strong predictive signal, while removing the RGB energy-field representation leaves aggregate trajectory error essentially unchanged. Increasing visual-backbone capacity also yields little benefit. The main scientific question is therefore not whether EVLA contains multiple modalities, but whether each modality changes the learned decision in the way intended by the benchmark construction.

\subsection{Modality Use versus Modality Availability}

The hidden regime is action-relevant inside the target generator, yet the learned model gains no aggregate accuracy from receiving the image that carries that regime. This separates two ideas that are often treated as equivalent in multimodal systems: information can be present in a modality without being exploited by the predictor.

Several mechanisms could produce this result. The regime-induced effect may be smaller than the language-conditioned variation; the texture may be difficult for the visual encoder to decode; the training subset may provide insufficient paired coverage; or the global MSE objective may reward dominant average structure while underweighting regime-specific changes. The present ablation does not identify which mechanism dominates.

A decisive visual-context experiment would keep the base state and processed language fixed, vary only \(Z\), and compare the resulting prediction shift with the reference-generator shift. It should also use a common candidate-action space across regimes so that the intervention tests contextual sensitivity rather than changes in the generator's available actions. This paired design is more informative for the visual pathway than another comparison of image backbones with nearly identical aggregate errors.

\subsection{Processed Language and Objective Semantics}

The language result is stronger within the implemented multimodal pathway, but
its interpretation depends on the executed loader. The dataset stores five
objective classes and associated instructions, whereas the training pipeline
reconstructs an internal objective using keyword rules, samples a new paraphrase,
and supplies the inferred objective to the deterministic baseline. Processed text
and objective-conditioned baseline information are therefore related but not
identical channels.

The large degradation observed when the text encoder is removed shows that this
multimodal predictor makes substantial use of the processed language
representation. It does not establish that language is necessary for low
aggregate error across all model families: the numerical-only MLP and LSTM
baselines achieve MSE values close to the designated MobileNet entry. Nor does
the current experiment establish semantic generalization across the five stored
objectives. Those stronger questions require a single objective ontology,
held-out paraphrases, and interventions that change both the text and every
objective-conditioned component together.

The text-scrambling diagnostic reinforces the value of this distinction. Models with similar nominal trajectory error can respond very differently when the text is disrupted. Language sensitivity is therefore a useful model-selection criterion alongside MSE, particularly for an intent-conditioned benchmark.

\subsection{Architecture and Model Selection}

The architecture study shows that greater structural complexity is not automatically rewarded by the current loss. Simpler learned variants achieve errors comparable to, or lower than, the full FiLM--Transformer model, and heavier visual encoders provide no clear accuracy advantage. At the same time, the no-residual model is much less sensitive to text scrambling than the full FiLM and MLP-fusion variants.

For EVLA, model selection should consequently balance trajectory accuracy, modality sensitivity, computational cost, and stability across seeds. Selecting only the lowest aggregate MSE would favor models that reproduce dominant target structure even if they underuse the intended conditioning signal. This is a general issue for multimodal decision models: predictive fit and conditional responsiveness are related, but they are not interchangeable.

\subsection{Implications for Residential Energy AI}

EVLA extends the VLA abstraction from embodied robotics to a residential energy setting in which the action is a continuous battery trajectory and the intent is supplied through language. The benchmark is useful precisely because its counterfactual organization makes modality dependence inspectable. The initial experiments also show why that inspection is necessary: adding a modality to the architecture does not guarantee that it affects the learned decision.

The next stage of the benchmark should therefore emphasize intervention quality rather than architectural breadth. Group-aware splits, held-out households and paraphrases, paired regime tests, and a cleaner objective pipeline would provide stronger evidence about generalization and conditional behavior. The physical assumptions and reproducibility issues that bound the current study are detailed next.
\section{Limitations and Future Work}
\label{sec:limitations}

The current release is intentionally controlled, but several implementation choices limit the conclusions that can be drawn from it. These limitations concern the target generator, data partitioning, modality interventions, language processing, and experimental reproducibility.

\subsection{Sampling-Based Reference Generator and Action Semantics}

The sampling-based reference trajectory generator uses heuristic state updates, regime-conditioned candidate-action distributions, and manually specified objective costs. Its temperature proxy is not derived from an identified HVAC or building-envelope model, and the stored action variable is a normalized benchmark quantity whose physical power unit is not independently validated. The reference trajectory \(A^\star\) should therefore be read as the optimum among sampled candidates under the stated benchmark equations.

This distinction matters for the operational diagnostics. MSE, MAE, and RMSE are meaningful in the stored-action scale, while energy-cost, action-throughput, self-consumption, and SOC-bound quantities are internal diagnostics tied to the same scale and heuristic dynamics. A deployment-oriented extension would require calibrated battery and thermal models, explicit physical units, validated tariffs and forecasts, closed-loop feasibility checks, and a clearly defined mapping from predicted action to residential equipment.

\subsection{Experimental Scale and Data Partitioning}

The reported models are trained on a deliberately limited experimental subset
of 5,000 instances and evaluated on 500 validation and 500 test instances. This
subset was used to make the initial 36-configuration, three-seed study
computationally tractable and should be viewed as a pilot-scale evaluation of
the generated benchmark rather than as training on the full 6.588-million
instance collection. The implemented split groups files by household directory,
sorts filenames lexicographically, and applies fixed percentage boundaries. It
does not explicitly keep all 15 counterfactual variants of each base window
together, verify physical timestamps, or purge overlapping source windows.

Accordingly, the present numerical results are descriptive of this fixed subset.
The limited sample size itself is not treated as a correctness failure; the
stronger methodological limitation is that the partition is not guaranteed to
be base-group-disjoint or overlap-free. A stronger protocol should partition by
base-state identifier, keep all related \((Z,L)\) variants in the same split,
introduce a purge interval longer than any source-window overlap, and publish the
exact split manifests. Learning curves over progressively larger group-aware
subsets would then show whether the observed modality patterns persist with
scale. Household-held-out evaluation should be added before claims about
transfer beyond the residential records represented during training.

\subsection{Visual Representation, Preprocessing, and Regime Intervention}

The RGB energy field is a designed representation of power, temporal activity, and regime texture. The executed loader independently min--max normalizes each image channel and does not apply the canonical ImageNet mean--standard-deviation transform, even when an ImageNet-pretrained backbone is used. The per-image normalization retains relative spatial patterns but can remove absolute cross-image amplitude information, while the ImageNet mismatch may affect transfer from pretrained visual features. The present experiments do not isolate either preprocessing effect. A focused future ablation should compare the executed preprocessing against (i) fixed dataset-level scaling that preserves cross-image amplitude and (ii) canonical ImageNet normalization for pretrained backbones, using matched splits and seeds.

The current no-vision ablation shows that this representation is not required to achieve the reported aggregate error, but aggregate MSE alone cannot determine whether the model responds correctly to a regime change.

For a fixed state \(S\) and language objective \(L\), a paired regime test can compare
\begin{equation}
\begin{aligned}
    \Delta^{\star}_{z,z'}
    &= A^\star(S,z,L)-A^\star(S,z^\prime,L),\\
    \widehat{\Delta}_{z,z'}
    &= \hat{A}(S,V_z,L)-\hat{A}(S,V_{z'},L).
\end{aligned}
\end{equation}
The direction and magnitude of \(\widehat{\Delta}_{z,z'}\) should be evaluated against \(\Delta^{\star}_{z,z'}\) while all other inputs remain fixed. Future versions of the reference trajectory generator should also use a common candidate-action space across regimes. Comparisons with direct time-series encoders, temporal convolutional networks, recurrent models, and time-series Transformers would test whether the image representation itself offers value beyond the controlled intervention mechanism.

\subsection{Language Ontology and Generalization}

The generated dataset contains five objective classes with multiple paraphrases, but the executed loader reconstructs one of six internal objectives from keywords and may replace the stored instruction with a new paraphrase. Several original instructions consequently map to the same internal objective, and the available paraphrases are not partitioned into disjoint train and test sets.

Future experiments should preserve a single objective ontology from dataset generation through evaluation and reserve complete paraphrases for testing. The evaluation should include unseen formulations of known objectives, human-written instructions, ambiguous or incomplete requests, conflicting preferences, compositional objectives, and instructions outside the training ontology. A language intervention should modify both the text and every objective-conditioned component of the pipeline so that the measured response corresponds to a complete change in intent.

\subsection{Robustness, Training, and Reproducibility}

The experimental configuration contains disturbance-related fields, but the loader does not apply the associated perturbations; those labels therefore cannot support a robustness claim. Future stress tests should implement traceable perturbations to numerical inputs, images, language, tariffs, and forecasts and apply the same sampled perturbations to every compared model.

The executed language-conditioned smoothness rule is also mini-batch dependent:
if any internal objective label in a batch contains ``peak,'' the global
smoothness term for that whole batch is multiplied by 1.5. This means that
non-peak samples in the same batch inherit the increased smoothness weight. A
cleaner future implementation should compute this weighting per sample before
aggregation, or remove the objective-dependent multiplier entirely.

The statistical study uses three training seeds and includes some repeated
effective architectures under different labels. Random-number generators are
initialized at the seed-level run rather than immediately before every
configuration, so equivalent configurations need not share matched
initializations or data order. The training loop also saves the
best-validation checkpoint without reloading it before the final test pass.
These choices do not invalidate the reported descriptive results, but they make
fine-grained ranking among configurations with nearly identical errors
unreliable. Future evaluations should use matched seeds for paired ablations,
reload the selected checkpoint, increase the number of independent runs, and
report uncertainty measures appropriate to the comparison.

Bitwise reproduction also requires stable hashing, explicit initialization of every random-number generator, a fixed \texttt{PYTHONHASHSEED}, immutable split manifests, and checksums for generated samples, checkpoints, and result files. Inference timing should be measured after warm-up with explicit device synchronization and a stated batch size, precision mode, hardware configuration, and summary statistic.

\subsection{Generalization Beyond the Current Households}

EVLA is derived from processed records associated with 19 household files. Generalization to unseen households, climates, tariffs, building types, occupant behavior, battery capacities, and appliance compositions remains untested. Future benchmark versions should withhold complete households and broaden the operating conditions represented in training and evaluation. Extensions that coordinate battery operation with HVAC, photovoltaic generation, electric-vehicle charging, and flexible appliances would also move the task closer to realistic multi-device residential energy management.
\section{Reproducibility and Data Availability}
\label{sec:reproducibility}

The reproducibility materials associated with this study comprise the LaTeX manuscript, generated tables and figures, dataset-generation code, model-training and evaluation code, selected metadata summaries, dataset-audit outputs, and aggregated experimental results. These materials support inspection of the implemented pipeline and reconstruction of the manuscript tables from the supplied result files.

\subsection{Dataset and Audit Artifacts}

The dataset-summary artifacts preserve the scale and structural checks reported in Section~\ref{sec:results} together with additional audit fields. The stored metadata contain 7,496 distinct state-of-charge values and 6,645 distinct indoor-temperature values. The accompanying audit records JSON parseability, state-field presence, trajectory length, within-group metadata consistency, and the sampled image-availability check, allowing the reported dataset validation to be reconstructed from the release artifacts.

\subsection{Experimental Reproduction}

The experimental code records the subset-selection seed as 42 and uses model
seeds 42, 43, and 44 for the reported three-seed evaluation. The result
archive contains the seed-level and aggregated outputs used to construct the
reported ablation tables.

However, exact reproduction of the 5,000/500/500 subset requires the same
complete file collection, directory structure, and lexicographic filename
ordering. The prepared materials do not establish that a standalone manifest
of every selected training, validation, and test filename is available.
Accordingly, the subset should be described as deterministically selected by
the supplied script under a fixed file layout, rather than as independently
reconstructable from a separately released split manifest.

The software and hardware configuration is documented in the supplementary
material. The primary training configuration uses AdamW, a learning rate of
\(3\times10^{-4}\), weight decay 0.01, batch size 32, a maximum of 100 epochs,
cosine-annealing warm restarts, gradient clipping at 1.0, and early stopping
after 10 epochs without improvement in validation MSE.

\subsection{Determinism and Reproducibility Limitations}

The dataset-generation and evaluation pipelines initialize several random seeds, but they do not guarantee bitwise-identical reproduction across independent executions. In particular, parts of the generation code derive seeds using Python's built-in \texttt{hash} function. Because the result of this function may vary across interpreter processes unless \texttt{PYTHONHASHSEED} is fixed, nominally identical executions may produce different derived seeds. Paraphrase selection and some image transformations also rely on global random-number generators whose states must be initialized and recorded to reproduce an identical generated instance.

For a bitwise-reproducible archival release, the pipeline requires stable hashing in place of Python's process-dependent built-in \texttt{hash} function, explicit initialization of the Python, NumPy, and learning-framework random-number generators, and a recorded \texttt{PYTHONHASHSEED}. An immutable experiment manifest should additionally record source-record and generated-instance identifiers, train--validation--test assignments, configuration files, software versions, and checksums for generated JSON files, images, checkpoints, and result files.

A fully reproducible evaluation protocol also requires reloading the checkpoint selected by the validation criterion before the final test pass. The present timing values are reported only as approximate within-platform diagnostics because the executed timing code does not include warm-up iterations or explicit device synchronization; a deployment-oriented latency study would require synchronized repeated measurements with the batch size, hardware, precision mode, and summary statistic fixed in advance.

\subsection{Data and Code Availability}

The residential electrical-load measurements used as the source time series are
the public REFIT dataset and cleaned REFIT release
\citep{murray2017refit,murray2016refitcleaned}. The generated EVLA
collection, code, configurations, and experimental results are separate research
artifacts derived from those public source records.

The public project website associated with EVLA is available at
\begin{center}
    \url{https://lyessaadsaoud.github.io/EVLA/}.
\end{center}

The project website provides information about the benchmark and its
reproducibility resources. The source-code repository and the complete
\(6{,}588{,}045\)-instance generated dataset are not publicly accessible at the
time of this preprint. The supplementary material provides the complete
configuration-level and seed-level result tables, dataset-validation summary,
and computing environment used for the reported study.

A future public release of the generated dataset and source code will be linked
through the project website and will include the corresponding release version,
license, access conditions, and integrity information where applicable. No
unverified DOI or archival dataset identifier is claimed in this manuscript.

At the time of this preprint, third parties can reproduce the benchmark-generation
methodology and reported analysis to the extent supported by the public REFIT
source data, the methodological specifications provided in the manuscript, and
the accompanying supplementary material.

\section{Conclusion}
\label{sec:conclusion}

This paper introduced EVLA, a controlled multimodal benchmark for intent-conditioned residential battery-trajectory prediction. EVLA combines a synthetic energy-field observation, a numerical operating state, and a natural-language objective with 16-step trajectories generated by a finite-horizon sampling-based reference trajectory generator. Its paired factorial organization expands each retained residential window across hidden regimes and language objectives, creating related examples with the same stored operating-state metadata while benchmark factors and their generated targets change by design.

The initial study exposes a clear imbalance in modality use within the
implemented multimodal architecture. Removing processed language from the
MobileNet-family pathway causes a large error increase, whereas removing the RGB
pathway does not improve aggregate trajectory error and larger visual backbones
provide little benefit. At the same time, numerical-only MLP and LSTM baselines
remain close to the designated MobileNet result, showing that the language
ablation is a statement about dependence of this architecture on its processed
language route rather than universal language necessity. Architecture experiments
further show that low prediction error alone is insufficient for an
intent-conditioned model: configurations with similar nominal accuracy can
differ sharply in their sensitivity to the text input.

These findings make modality intervention, rather than model size, the central next step for EVLA. Group-aware and purged splits, a consistent language ontology, held-out households and paraphrases, paired regime tests, validated action semantics, and stronger reproducibility controls will determine whether the benchmark can support reliable conclusions about conditional behavior and generalization. In this form, EVLA provides a concrete basis for studying how multimodal AI models translate residential state, semantic intent, and latent context into continuous energy-action trajectories.
\section*{Declaration of competing interest}

The authors declare that they have no known competing financial interests or
personal relationships that could have appeared to influence the work reported
in this paper.

\section*{Declaration of generative AI and AI-assisted technologies in the manuscript preparation process}

During manuscript preparation, AI-assisted tools were used for language editing.

\clearpage
\FloatBarrier

% ============================================================
% Supplementary Material appended to the archival preprint
% ============================================================
\setcounter{section}{0}
\setcounter{subsection}{0}
\setcounter{table}{0}
\setcounter{figure}{0}
\setcounter{equation}{0}
\renewcommand{\thesection}{S\arabic{section}}
\renewcommand{\thesubsection}{S\arabic{section}.\arabic{subsection}}
\renewcommand{\thetable}{S\arabic{table}}
\renewcommand{\thefigure}{S\arabic{figure}}
\renewcommand{\theequation}{S\arabic{equation}}

\begin{center}
{\Large\bfseries Supplementary Material}\par
\vspace{0.4em}
{\large\bfseries Energy Vision--Language--Action: A Controlled Multimodal Benchmark for Intent-Conditioned Residential Energy Management}\par
\vspace{0.5em}
Lyes Saad Saoud, Oualid Doukhi, Ehsan Reihani, Saeed Sepasi, Deok Jin Lee, Moussa Ayyash, and Reza Ghorbani
\end{center}
\vspace{1em}

\section{Purpose of the Supplementary Material}
\label{supp:purpose}

This supplementary material provides additional information that supports the EVLA manuscript without interrupting the main paper. The main manuscript presents the benchmark formulation, dataset construction, trajectory-prediction model, experimental protocol, and principal findings. This document provides complementary details on dataset validation, counterfactual organization, reference-trajectory construction, complete ablation tables, and the computational environment.

All information reported here refers to the same EVLA benchmark described in the main manuscript. No separate dataset version is introduced in this supplementary material.

\section{Detailed Dataset Validation}
\label{supp:dataset_validation}

The EVLA benchmark contains 439,203 retained base operating states and 6,588,045 multimodal samples. Each base operating state is expanded into 15 controlled variants formed by three hidden visual regimes and five language objectives. Table~\ref{tab:supp_validation} provides the detailed validation summary.

\begin{table}[t]
\centering
\caption{Detailed EVLA dataset validation summary.}
\label{tab:supp_validation}
\begin{tabular}{lr}
\toprule
Metric & Value \\
\midrule
Total multimodal samples & 6,588,045 \\
Unique base operating states & 439,203 \\
Expansion factor & 15 \\
Samples per \((Z,L)\) cell & 439,203 \\
Malformed JSON files & 0 \\
Missing state files & 0 \\
Trajectories with length not equal to 16 & 0 \\
Groups with inconsistent SOC or indoor temperature & 0 \\
Unique SOC values & 7,496 \\
Unique indoor-temperature values & 6,645 \\
Sampled images checked & 6,588 \\
Missing sampled images & 0 \\
\bottomrule
\end{tabular}
\end{table}

The validation confirms that related samples sharing the same base operating state preserve the stored physical-state metadata while the benchmark factors \(Z\) and \(L\) vary. Those factors also affect target generation by design. The grouping supports paired diagnostics, but it does not by itself make the executed train--validation--test partition leakage-controlled; the partitioning limitation is stated in the main manuscript.

\section{Paired-Factor Organization}
\label{supp:counterfactual_organization}

EVLA is organized around groups of samples that share the same explicit physical state. For each retained base operating state, the numerical state \(S\) is fixed, while the hidden visual regime \(Z\in\{0,1,2\}\) and language objective \(L\) are varied independently. This produces 15 samples per base state:
\[
    3 \ \text{hidden visual regimes}
    \times
    5 \ \text{language objectives}
    =
    15 \ \text{controlled variants}.
\]
The purpose of this design is to separate physical-state dependence from modality dependence. A model cannot infer the language objective from the physical state alone because language is sampled independently. Similarly, hidden-regime information is not given as a numerical input to the learned model; it is available only through the energy-field image.

\section{Supplementary Reference-Trajectory Generator Details}
\label{supp:reference_generator_details}

The sampling-based reference trajectory generator produces a 16-step battery-action trajectory for each tuple \((S,Z,L)\). Candidate action sequences are sampled, rolled out under regime-dependent dynamics, and scored using objective-dependent costs together with the implemented state-of-charge handling and smoothness term; the lowest-cost sampled sequence is stored as the reference trajectory. The five language-objective classes induce different preferences:
\begin{itemize}[leftmargin=*]
    \item \textbf{Energy saving:} favors cost-reducing actions while respecting state-of-charge limits and smoothness.
    \item \textbf{Comfort:} favors actions that support indoor-condition regulation and reduce thermal discomfort.
    \item \textbf{Balanced operation:} combines energy cost, comfort, battery cycling, and action smoothness without strongly prioritizing a single term.
    \item \textbf{Grid support:} favors actions that reduce stress during grid-support periods and shift energy use away from unfavorable operating windows.
    \item \textbf{Eco operation:} favors low-impact operation while maintaining feasibility and avoiding excessive cycling.
\end{itemize}

These scenarios are not intended to represent certified closed-loop building control. Their role is to provide a controlled sampling-based reference target for multimodal trajectory-prediction experiments.

\section{Complementary Qualitative Diagnostics}
\label{supp:qualitative_diagnostics}

The main manuscript reports the principal dataset figures. This section provides additional qualitative diagnostics to help inspect the relationship between energy-field observations and reference action trajectories.

\begin{figure}[t]
\centering
\includegraphics[width=0.95\linewidth]{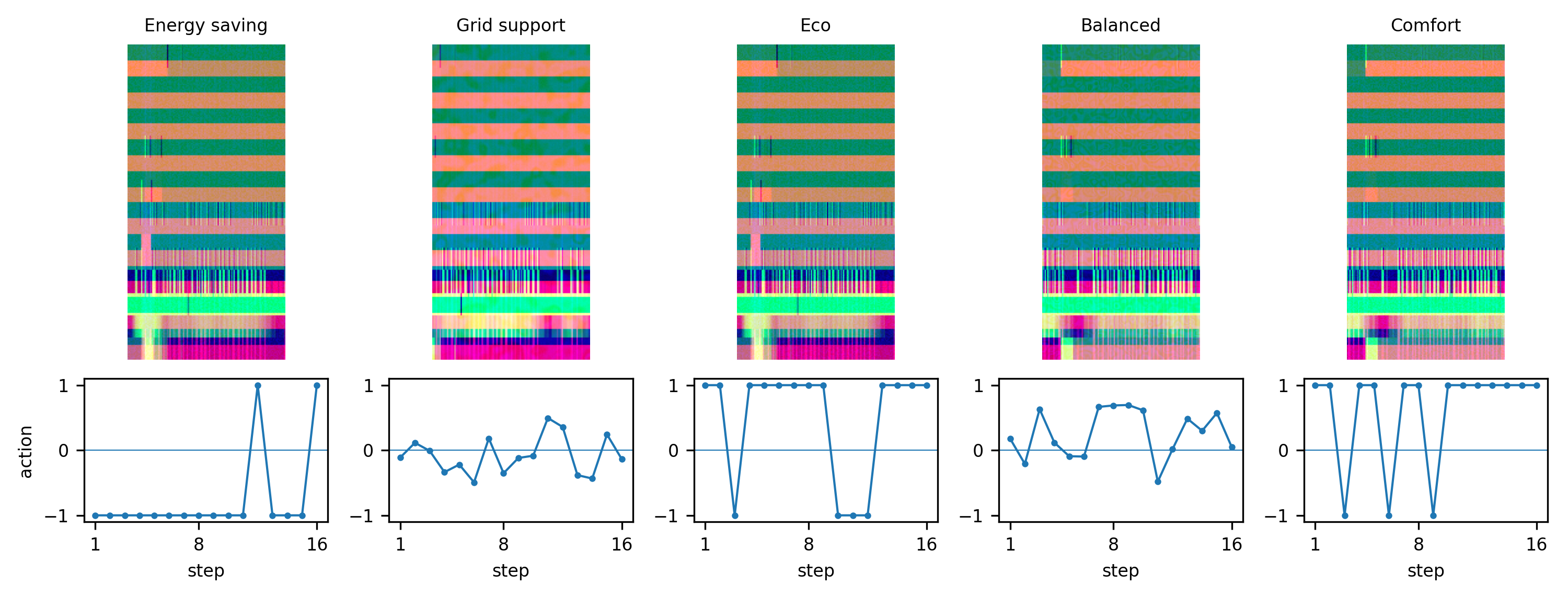}
\caption{Representative EVLA energy-field observations and corresponding 16-step action trajectories.}
\label{fig:supp_energy_fields_actions}
\end{figure}

\begin{figure}[t]
\centering
\includegraphics[width=0.95\linewidth]{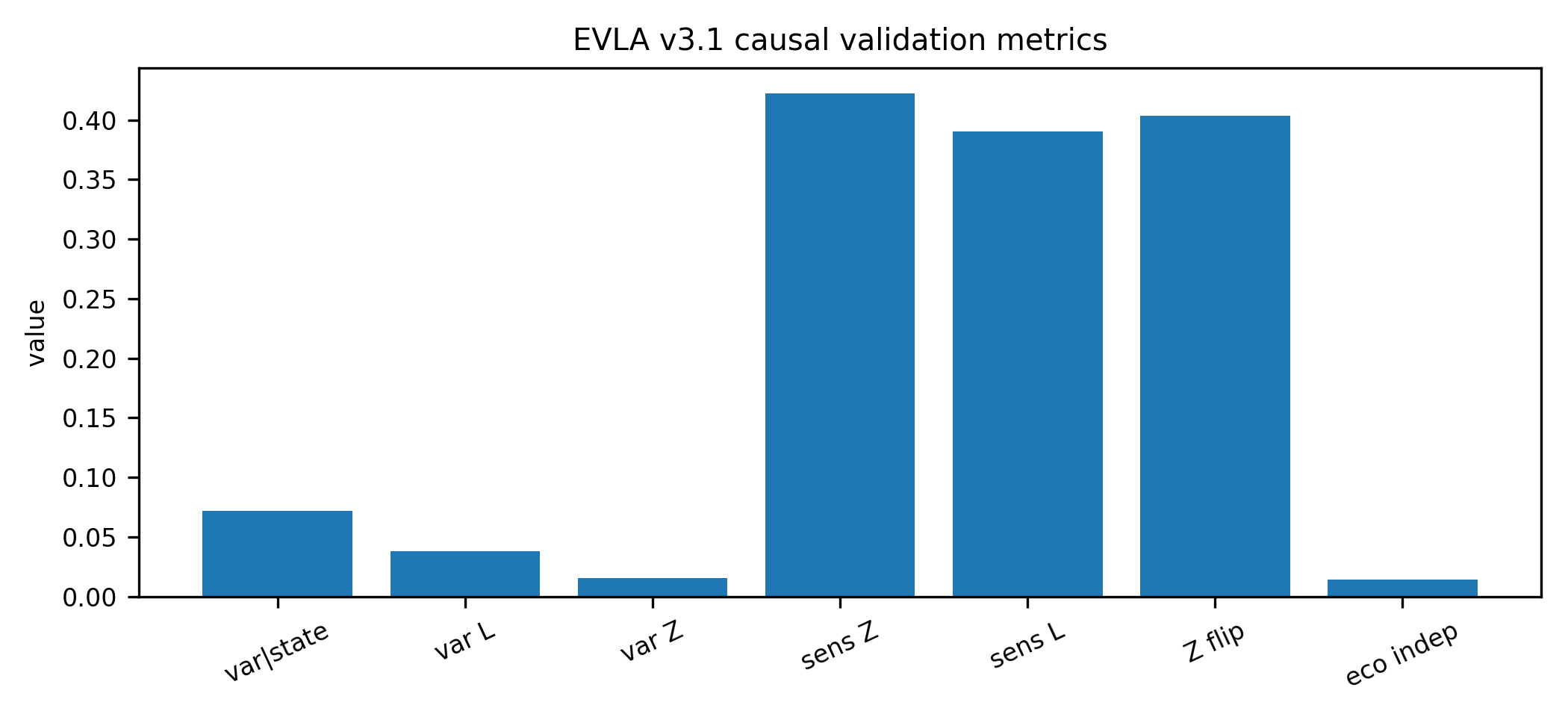}
\caption{Diagnostic used to inspect the paired multimodal construction.}
\label{fig:supp_causal_validation}
\end{figure}

\section{Complete Ablation Results}
\label{supp:complete_ablation_tables}

The main manuscript reports compact ablation summaries and discusses the
principal trends. The following tables provide the full configuration-level
and seed-level results used to obtain those summaries; no superseded draft
tables or archived result bundles are reproduced here.

\clearpage
\section{Complete Configuration-Level Ablation Results}
\label{supp:complete_36}

Table~\ref{tab:supp_36_full} reports the complete 36-configuration aggregation
used for the main-manuscript ablation summaries. Values are mean
\(\pm\) sample standard deviation over seeds 42, 43, and 44.

\begin{landscape}
\scriptsize
\setlength{\tabcolsep}{3.2pt}
\begin{longtable}{@{}lccccccc@{}}
\caption{Complete 36-configuration ablation results aggregated over seeds 42, 43, and 44.}
\label{tab:supp_36_full}\\
\toprule
Variant & MSE & MAE & RMSE & Params (M) & Inf. (ms) & Energy & Viol. \\
\midrule
\endfirsthead
\multicolumn{8}{c}{\tablename\ \thetable\ -- continued from previous page}\\
\toprule
Variant & MSE & MAE & RMSE & Params (M) & Inf. (ms) & Energy & Viol. \\
\midrule
\endhead
\midrule
\multicolumn{8}{r}{Continued on next page}\\
\endfoot
\bottomrule
\endlastfoot
\texttt{Backbone-Mobilenet} & 0.3829$\pm$0.0003 & 0.4850$\pm$0.0007 & 0.6188$\pm$0.0002 & 4.44$\pm$0.00 & 0.495$\pm$0.004 & 0.0347$\pm$0.0035 & 0.0000$\pm$0.0000 \\
\texttt{Arch-NoResidual-MobileNet} & 0.3829$\pm$0.0001 & 0.4851$\pm$0.0001 & 0.6188$\pm$0.0001 & 4.44$\pm$0.00 & 0.490$\pm$0.003 & 0.0392$\pm$0.0002 & 0.0000$\pm$0.0000 \\
\texttt{Arch-MLPFusion-MobileNet} & 0.3832$\pm$0.0010 & 0.4856$\pm$0.0018 & 0.6190$\pm$0.0008 & 1.46$\pm$0.00 & 0.453$\pm$0.014 & 0.0382$\pm$0.0063 & 0.0000$\pm$0.0000 \\
\texttt{Modality-NoVision-Resnet} & 0.3833$\pm$0.0006 & 0.4857$\pm$0.0004 & 0.6191$\pm$0.0005 & 3.33$\pm$0.00 & 0.186$\pm$0.007 & 0.0384$\pm$0.0011 & 0.0000$\pm$0.0000 \\
\texttt{Modality-Full-Convnext} & 0.3834$\pm$0.0010 & 0.4857$\pm$0.0012 & 0.6192$\pm$0.0008 & 19.36$\pm$0.00 & 1.289$\pm$0.013 & 0.0391$\pm$0.0038 & 0.0000$\pm$0.0000 \\
\texttt{Backbone-Simclr} & 0.3834$\pm$0.0005 & 0.4856$\pm$0.0005 & 0.6192$\pm$0.0004 & 14.93$\pm$0.00 & 0.788$\pm$0.002 & 0.0377$\pm$0.0017 & 0.0000$\pm$0.0000 \\
\texttt{Backbone-Resnet} & 0.3835$\pm$0.0004 & 0.4857$\pm$0.0007 & 0.6193$\pm$0.0003 & 14.26$\pm$0.00 & 0.823$\pm$0.005 & 0.0386$\pm$0.0024 & 0.0000$\pm$0.0000 \\
\texttt{Modality-NoVision-Swin} & 0.3836$\pm$0.0003 & 0.4860$\pm$0.0006 & 0.6194$\pm$0.0003 & 3.33$\pm$0.00 & 0.187$\pm$0.003 & 0.0388$\pm$0.0021 & 0.0000$\pm$0.0000 \\
\texttt{Modality-NoVision-Simclr} & 0.3837$\pm$0.0003 & 0.4861$\pm$0.0004 & 0.6194$\pm$0.0002 & 3.33$\pm$0.00 & 0.189$\pm$0.005 & 0.0368$\pm$0.0041 & 0.0000$\pm$0.0000 \\
\texttt{Modality-Full-Swin} & 0.3840$\pm$0.0004 & 0.4865$\pm$0.0007 & 0.6197$\pm$0.0004 & 19.26$\pm$0.00 & 1.244$\pm$0.002 & 0.0400$\pm$0.0023 & 0.0000$\pm$0.0000 \\
\texttt{Modality-NoVision-Tinycnn} & 0.3841$\pm$0.0008 & 0.4865$\pm$0.0010 & 0.6198$\pm$0.0007 & 3.33$\pm$0.00 & 0.194$\pm$0.011 & 0.0405$\pm$0.0037 & 0.0000$\pm$0.0000 \\
\texttt{Modality-NoVision-Convnext} & 0.3843$\pm$0.0015 & 0.4868$\pm$0.0015 & 0.6199$\pm$0.0012 & 3.33$\pm$0.00 & 0.190$\pm$0.005 & 0.0426$\pm$0.0047 & 0.0000$\pm$0.0000 \\
\texttt{Modality-NoVision-Mobilenet} & 0.3843$\pm$0.0013 & 0.4867$\pm$0.0018 & 0.6199$\pm$0.0010 & 3.33$\pm$0.00 & 0.191$\pm$0.010 & 0.0397$\pm$0.0099 & 0.0000$\pm$0.0000 \\
\texttt{Modality-Full-Resnet} & 0.3846$\pm$0.0016 & 0.4870$\pm$0.0015 & 0.6202$\pm$0.0013 & 14.26$\pm$0.00 & 0.826$\pm$0.004 & 0.0423$\pm$0.0045 & 0.0000$\pm$0.0000 \\
\texttt{Modality-Full-Simclr} & 0.3846$\pm$0.0014 & 0.4872$\pm$0.0017 & 0.6202$\pm$0.0012 & 14.93$\pm$0.00 & 0.781$\pm$0.006 & 0.0423$\pm$0.0050 & 0.0000$\pm$0.0000 \\
\texttt{Backbone-Tinycnn} & 0.3846$\pm$0.0013 & 0.4872$\pm$0.0016 & 0.6202$\pm$0.0010 & 4.17$\pm$0.00 & 0.298$\pm$0.005 & 0.0421$\pm$0.0050 & 0.0000$\pm$0.0000 \\
\texttt{Backbone-Convnext} & 0.3849$\pm$0.0019 & 0.4872$\pm$0.0021 & 0.6204$\pm$0.0015 & 19.36$\pm$0.00 & 1.282$\pm$0.002 & 0.0421$\pm$0.0061 & 0.0000$\pm$0.0000 \\
\texttt{Backbone-Swin} & 0.3854$\pm$0.0024 & 0.4878$\pm$0.0030 & 0.6208$\pm$0.0019 & 19.26$\pm$0.00 & 1.236$\pm$0.007 & 0.0435$\pm$0.0097 & 0.0000$\pm$0.0000 \\
\texttt{Modality-Full-Mobilenet} & 0.3856$\pm$0.0039 & 0.4879$\pm$0.0043 & 0.6210$\pm$0.0032 & 4.44$\pm$0.00 & 0.483$\pm$0.005 & 0.0422$\pm$0.0131 & 0.0000$\pm$0.0000 \\
\texttt{Baseline-MLP} & 0.3887$\pm$0.0001 & 0.4905$\pm$0.0001 & 0.6234$\pm$0.0001 & 0.07$\pm$0.00 & 0.008$\pm$0.001 & 0.0402$\pm$0.0004 & 0.0000$\pm$0.0000 \\
\texttt{Baseline-LSTM} & 0.3893$\pm$0.0004 & 0.4913$\pm$0.0005 & 0.6240$\pm$0.0003 & 1.06$\pm$0.00 & 0.134$\pm$0.002 & 0.0423$\pm$0.0017 & 0.0000$\pm$0.0000 \\
\texttt{Modality-Full-Tinycnn} & 0.3904$\pm$0.0119 & 0.4910$\pm$0.0090 & 0.6248$\pm$0.0095 & 4.17$\pm$0.00 & 0.297$\pm$0.006 & 0.0407$\pm$0.0057 & 0.0000$\pm$0.0000 \\
\texttt{Arch-Full-MobileNet} & 0.3922$\pm$0.0111 & 0.4928$\pm$0.0082 & 0.6262$\pm$0.0089 & 4.44$\pm$0.00 & 0.490$\pm$0.011 & 0.0466$\pm$0.0075 & 0.0000$\pm$0.0000 \\
\texttt{Modality-StateOnly-Convnext} & 0.8626$\pm$0.0000 & 0.7190$\pm$0.0011 & 0.9288$\pm$0.0000 & 3.18$\pm$0.00 & 0.173$\pm$0.006 & 0.0882$\pm$0.0004 & 0.0023$\pm$0.0000 \\
\texttt{Modality-StateOnly-Mobilenet} & 0.8626$\pm$0.0000 & 0.7188$\pm$0.0017 & 0.9288$\pm$0.0000 & 3.18$\pm$0.00 & 0.171$\pm$0.015 & 0.0882$\pm$0.0005 & 0.0023$\pm$0.0000 \\
\texttt{Modality-NoLanguage-Resnet} & 0.8627$\pm$0.0001 & 0.7208$\pm$0.0010 & 0.9288$\pm$0.0001 & 14.11$\pm$0.00 & 0.802$\pm$0.001 & 0.0877$\pm$0.0003 & 0.0022$\pm$0.0000 \\
\texttt{Modality-NoLanguage-Simclr} & 0.8627$\pm$0.0001 & 0.7203$\pm$0.0018 & 0.9288$\pm$0.0001 & 14.78$\pm$0.00 & 0.763$\pm$0.004 & 0.0878$\pm$0.0005 & 0.0022$\pm$0.0000 \\
\texttt{Modality-NoLanguage-Convnext} & 0.8627$\pm$0.0000 & 0.7202$\pm$0.0015 & 0.9288$\pm$0.0000 & 19.22$\pm$0.00 & 1.273$\pm$0.004 & 0.0878$\pm$0.0004 & 0.0022$\pm$0.0000 \\
\texttt{Modality-StateOnly-Resnet} & 0.8627$\pm$0.0000 & 0.7189$\pm$0.0023 & 0.9288$\pm$0.0000 & 3.18$\pm$0.00 & 0.177$\pm$0.008 & 0.0882$\pm$0.0007 & 0.0023$\pm$0.0000 \\
\texttt{Modality-StateOnly-Swin} & 0.8627$\pm$0.0001 & 0.7198$\pm$0.0024 & 0.9288$\pm$0.0001 & 3.18$\pm$0.00 & 0.183$\pm$0.005 & 0.0880$\pm$0.0007 & 0.0022$\pm$0.0000 \\
\texttt{Modality-NoLanguage-Mobilenet} & 0.8628$\pm$0.0001 & 0.7211$\pm$0.0012 & 0.9288$\pm$0.0001 & 4.29$\pm$0.00 & 0.485$\pm$0.005 & 0.0876$\pm$0.0003 & 0.0022$\pm$0.0000 \\
\texttt{Modality-StateOnly-Simclr} & 0.8628$\pm$0.0001 & 0.7170$\pm$0.0019 & 0.9289$\pm$0.0001 & 3.18$\pm$0.00 & 0.184$\pm$0.003 & 0.0888$\pm$0.0006 & 0.0023$\pm$0.0000 \\
\texttt{Modality-NoLanguage-Swin} & 0.8629$\pm$0.0004 & 0.7172$\pm$0.0051 & 0.9289$\pm$0.0002 & 19.11$\pm$0.00 & 1.226$\pm$0.004 & 0.0888$\pm$0.0016 & 0.0023$\pm$0.0001 \\
\texttt{Modality-StateOnly-Tinycnn} & 0.8630$\pm$0.0003 & 0.7193$\pm$0.0048 & 0.9290$\pm$0.0001 & 3.18$\pm$0.00 & 0.179$\pm$0.003 & 0.0882$\pm$0.0015 & 0.0023$\pm$0.0001 \\
\texttt{Modality-NoLanguage-Tinycnn} & 0.8630$\pm$0.0004 & 0.7201$\pm$0.0052 & 0.9290$\pm$0.0002 & 4.02$\pm$0.00 & 0.266$\pm$0.004 & 0.0879$\pm$0.0015 & 0.0022$\pm$0.0001 \\
\texttt{Arch-BaseOnly-MobileNet} & 8.0795$\pm$0.0000 & 2.6864$\pm$0.0000 & 2.8425$\pm$0.0000 & 1.26$\pm$0.00 & 0.114$\pm$0.005 & 1.0925$\pm$0.0000 & 0.1084$\pm$0.0000 \\
\end{longtable}
\end{landscape}

\clearpage
\section{Complete Seed-Level Results}
\label{supp:complete_108}

Table~\ref{tab:supp_seed_108_full} reports the complete seed-level record for
all 108 runs used to compute the configuration-level summaries.

\begin{landscape}
\scriptsize
\setlength{\tabcolsep}{3.5pt}
\begin{longtable}{@{}lrrrrrrr@{}}
\caption{Complete seed-level results for the 108 reported runs.}
\label{tab:supp_seed_108_full}\\
\toprule
Run & MSE & MAE & RMSE & Energy & Viol. & Inf. (ms) & Params (M) \\
\midrule
\endfirsthead
\multicolumn{8}{c}{\tablename\ \thetable\ -- continued from previous page}\\
\toprule
Run & MSE & MAE & RMSE & Energy & Viol. & Inf. (ms) & Params (M) \\
\midrule
\endhead
\midrule
\multicolumn{8}{r}{Continued on next page}\\
\endfoot
\bottomrule
\endlastfoot
\texttt{seed42-Arch-BaseOnly-MobileNet} & 8.0795 & 2.6864 & 2.8425 & 1.0925 & 0.1084 & 0.113 & 1.26 \\
\texttt{seed43-Arch-BaseOnly-MobileNet} & 8.0795 & 2.6864 & 2.8425 & 1.0925 & 0.1084 & 0.108 & 1.26 \\
\texttt{seed44-Arch-BaseOnly-MobileNet} & 8.0795 & 2.6864 & 2.8425 & 1.0925 & 0.1084 & 0.119 & 1.26 \\
\texttt{seed42-Arch-Full-MobileNet} & 0.4047 & 0.5017 & 0.6362 & 0.0496 & 0.0000 & 0.490 & 4.44 \\
\texttt{seed43-Arch-Full-MobileNet} & 0.3833 & 0.4856 & 0.6191 & 0.0381 & 0.0000 & 0.479 & 4.44 \\
\texttt{seed44-Arch-Full-MobileNet} & 0.3887 & 0.4911 & 0.6234 & 0.0521 & 0.0000 & 0.502 & 4.44 \\
\texttt{seed42-Arch-MLPFusion-MobileNet} & 0.3822 & 0.4838 & 0.6183 & 0.0314 & 0.0000 & 0.440 & 1.46 \\
\texttt{seed43-Arch-MLPFusion-MobileNet} & 0.3841 & 0.4874 & 0.6198 & 0.0437 & 0.0000 & 0.452 & 1.46 \\
\texttt{seed44-Arch-MLPFusion-MobileNet} & 0.3832 & 0.4856 & 0.6190 & 0.0395 & 0.0000 & 0.468 & 1.46 \\
\texttt{seed42-Arch-NoResidual-MobileNet} & 0.3830 & 0.4851 & 0.6188 & 0.0395 & 0.0000 & 0.492 & 4.44 \\
\texttt{seed43-Arch-NoResidual-MobileNet} & 0.3828 & 0.4850 & 0.6187 & 0.0391 & 0.0000 & 0.486 & 4.44 \\
\texttt{seed44-Arch-NoResidual-MobileNet} & 0.3830 & 0.4852 & 0.6189 & 0.0390 & 0.0000 & 0.491 & 4.44 \\
\texttt{seed42-Backbone-Convnext} & 0.3832 & 0.4854 & 0.6190 & 0.0380 & 0.0000 & 1.280 & 19.36 \\
\texttt{seed43-Backbone-Convnext} & 0.3845 & 0.4867 & 0.6201 & 0.0391 & 0.0000 & 1.283 & 19.36 \\
\texttt{seed44-Backbone-Convnext} & 0.3869 & 0.4895 & 0.6220 & 0.0492 & 0.0000 & 1.283 & 19.36 \\
\texttt{seed42-Backbone-Mobilenet} & 0.3828 & 0.4847 & 0.6187 & 0.0352 & 0.0000 & 0.500 & 4.44 \\
\texttt{seed43-Backbone-Mobilenet} & 0.3832 & 0.4858 & 0.6191 & 0.0379 & 0.0000 & 0.492 & 4.44 \\
\texttt{seed44-Backbone-Mobilenet} & 0.3827 & 0.4845 & 0.6186 & 0.0310 & 0.0000 & 0.493 & 4.44 \\
\texttt{seed42-Backbone-Resnet} & 0.3831 & 0.4851 & 0.6190 & 0.0364 & 0.0000 & 0.817 & 14.26 \\
\texttt{seed43-Backbone-Resnet} & 0.3834 & 0.4857 & 0.6192 & 0.0382 & 0.0000 & 0.825 & 14.26 \\
\texttt{seed44-Backbone-Resnet} & 0.3840 & 0.4864 & 0.6196 & 0.0412 & 0.0000 & 0.827 & 14.26 \\
\texttt{seed42-Backbone-Simclr} & 0.3838 & 0.4859 & 0.6195 & 0.0388 & 0.0000 & 0.788 & 14.93 \\
\texttt{seed43-Backbone-Simclr} & 0.3829 & 0.4850 & 0.6188 & 0.0357 & 0.0000 & 0.790 & 14.93 \\
\texttt{seed44-Backbone-Simclr} & 0.3835 & 0.4859 & 0.6193 & 0.0386 & 0.0000 & 0.786 & 14.93 \\
\texttt{seed42-Backbone-Swin} & 0.3858 & 0.4885 & 0.6211 & 0.0477 & 0.0000 & 1.235 & 19.26 \\
\texttt{seed43-Backbone-Swin} & 0.3876 & 0.4904 & 0.6226 & 0.0504 & 0.0000 & 1.229 & 19.26 \\
\texttt{seed44-Backbone-Swin} & 0.3828 & 0.4846 & 0.6187 & 0.0325 & 0.0000 & 1.244 & 19.26 \\
\texttt{seed42-Backbone-Tinycnn} & 0.3860 & 0.4890 & 0.6213 & 0.0478 & 0.0000 & 0.296 & 4.17 \\
\texttt{seed43-Backbone-Tinycnn} & 0.3844 & 0.4870 & 0.6200 & 0.0403 & 0.0000 & 0.294 & 4.17 \\
\texttt{seed44-Backbone-Tinycnn} & 0.3834 & 0.4857 & 0.6192 & 0.0383 & 0.0000 & 0.303 & 4.17 \\
\texttt{seed42-Baseline-LSTM} & 0.3898 & 0.4919 & 0.6243 & 0.0443 & 0.0000 & 0.132 & 1.06 \\
\texttt{seed43-Baseline-LSTM} & 0.3891 & 0.4911 & 0.6238 & 0.0414 & 0.0000 & 0.137 & 1.06 \\
\texttt{seed44-Baseline-LSTM} & 0.3891 & 0.4910 & 0.6238 & 0.0413 & 0.0000 & 0.133 & 1.06 \\
\texttt{seed42-Baseline-MLP} & 0.3888 & 0.4906 & 0.6235 & 0.0405 & 0.0000 & 0.010 & 0.07 \\
\texttt{seed43-Baseline-MLP} & 0.3887 & 0.4906 & 0.6235 & 0.0404 & 0.0000 & 0.007 & 0.07 \\
\texttt{seed44-Baseline-MLP} & 0.3886 & 0.4904 & 0.6234 & 0.0397 & 0.0000 & 0.008 & 0.07 \\
\texttt{seed42-Modality-Full-Convnext} & 0.3836 & 0.4858 & 0.6193 & 0.0402 & 0.0000 & 1.303 & 19.36 \\
\texttt{seed43-Modality-Full-Convnext} & 0.3843 & 0.4867 & 0.6199 & 0.0422 & 0.0000 & 1.281 & 19.36 \\
\texttt{seed44-Modality-Full-Convnext} & 0.3823 & 0.4844 & 0.6183 & 0.0348 & 0.0000 & 1.281 & 19.36 \\
\texttt{seed42-Modality-Full-Mobilenet} & 0.3828 & 0.4849 & 0.6187 & 0.0358 & 0.0000 & 0.479 & 4.44 \\
\texttt{seed43-Modality-Full-Mobilenet} & 0.3839 & 0.4858 & 0.6196 & 0.0335 & 0.0000 & 0.489 & 4.44 \\
\texttt{seed44-Modality-Full-Mobilenet} & 0.3901 & 0.4928 & 0.6246 & 0.0573 & 0.0000 & 0.482 & 4.44 \\
\texttt{seed42-Modality-Full-Resnet} & 0.3833 & 0.4856 & 0.6191 & 0.0379 & 0.0000 & 0.829 & 14.26 \\
\texttt{seed43-Modality-Full-Resnet} & 0.3843 & 0.4868 & 0.6199 & 0.0421 & 0.0000 & 0.822 & 14.26 \\
\texttt{seed44-Modality-Full-Resnet} & 0.3863 & 0.4887 & 0.6215 & 0.0469 & 0.0000 & 0.828 & 14.26 \\
\texttt{seed42-Modality-Full-Simclr} & 0.3837 & 0.4864 & 0.6195 & 0.0412 & 0.0000 & 0.774 & 14.93 \\
\texttt{seed43-Modality-Full-Simclr} & 0.3863 & 0.4891 & 0.6215 & 0.0477 & 0.0000 & 0.782 & 14.93 \\
\texttt{seed44-Modality-Full-Simclr} & 0.3839 & 0.4861 & 0.6196 & 0.0379 & 0.0000 & 0.786 & 14.93 \\
\texttt{seed42-Modality-Full-Swin} & 0.3837 & 0.4862 & 0.6195 & 0.0380 & 0.0000 & 1.245 & 19.26 \\
\texttt{seed43-Modality-Full-Swin} & 0.3845 & 0.4873 & 0.6201 & 0.0424 & 0.0000 & 1.241 & 19.26 \\
\texttt{seed44-Modality-Full-Swin} & 0.3837 & 0.4860 & 0.6194 & 0.0396 & 0.0000 & 1.245 & 19.26 \\
\texttt{seed42-Modality-Full-Tinycnn} & 0.4042 & 0.5013 & 0.6357 & 0.0467 & 0.0000 & 0.294 & 4.17 \\
\texttt{seed43-Modality-Full-Tinycnn} & 0.3836 & 0.4856 & 0.6193 & 0.0354 & 0.0000 & 0.293 & 4.17 \\
\texttt{seed44-Modality-Full-Tinycnn} & 0.3835 & 0.4861 & 0.6193 & 0.0400 & 0.0000 & 0.304 & 4.17 \\
\texttt{seed42-Modality-NoLanguage-Convnext} & 0.8627 & 0.7206 & 0.9288 & 0.0877 & 0.0022 & 1.271 & 19.22 \\
\texttt{seed43-Modality-NoLanguage-Convnext} & 0.8627 & 0.7186 & 0.9288 & 0.0883 & 0.0023 & 1.269 & 19.22 \\
\texttt{seed44-Modality-NoLanguage-Convnext} & 0.8628 & 0.7215 & 0.9288 & 0.0875 & 0.0022 & 1.278 & 19.22 \\
\texttt{seed42-Modality-NoLanguage-Mobilenet} & 0.8626 & 0.7197 & 0.9288 & 0.0880 & 0.0022 & 0.487 & 4.29 \\
\texttt{seed43-Modality-NoLanguage-Mobilenet} & 0.8629 & 0.7220 & 0.9289 & 0.0873 & 0.0022 & 0.488 & 4.29 \\
\texttt{seed44-Modality-NoLanguage-Mobilenet} & 0.8628 & 0.7215 & 0.9289 & 0.0874 & 0.0022 & 0.478 & 4.29 \\
\texttt{seed42-Modality-NoLanguage-Resnet} & 0.8625 & 0.7199 & 0.9287 & 0.0879 & 0.0022 & 0.803 & 14.11 \\
\texttt{seed43-Modality-NoLanguage-Resnet} & 0.8628 & 0.7218 & 0.9289 & 0.0874 & 0.0022 & 0.801 & 14.11 \\
\texttt{seed44-Modality-NoLanguage-Resnet} & 0.8627 & 0.7206 & 0.9288 & 0.0877 & 0.0022 & 0.801 & 14.11 \\
\texttt{seed42-Modality-NoLanguage-Simclr} & 0.8627 & 0.7199 & 0.9288 & 0.0879 & 0.0022 & 0.759 & 14.78 \\
\texttt{seed43-Modality-NoLanguage-Simclr} & 0.8626 & 0.7188 & 0.9288 & 0.0883 & 0.0023 & 0.762 & 14.78 \\
\texttt{seed44-Modality-NoLanguage-Simclr} & 0.8628 & 0.7223 & 0.9289 & 0.0872 & 0.0022 & 0.768 & 14.78 \\
\texttt{seed42-Modality-NoLanguage-Swin} & 0.8634 & 0.7116 & 0.9292 & 0.0906 & 0.0024 & 1.225 & 19.11 \\
\texttt{seed43-Modality-NoLanguage-Swin} & 0.8627 & 0.7185 & 0.9288 & 0.0884 & 0.0023 & 1.224 & 19.11 \\
\texttt{seed44-Modality-NoLanguage-Swin} & 0.8628 & 0.7216 & 0.9289 & 0.0874 & 0.0022 & 1.230 & 19.11 \\
\texttt{seed42-Modality-NoLanguage-Tinycnn} & 0.8627 & 0.7190 & 0.9288 & 0.0882 & 0.0023 & 0.262 & 4.02 \\
\texttt{seed43-Modality-NoLanguage-Tinycnn} & 0.8635 & 0.7257 & 0.9292 & 0.0863 & 0.0021 & 0.266 & 4.02 \\
\texttt{seed44-Modality-NoLanguage-Tinycnn} & 0.8630 & 0.7155 & 0.9290 & 0.0893 & 0.0023 & 0.271 & 4.02 \\
\texttt{seed42-Modality-NoVision-Convnext} & 0.3860 & 0.4885 & 0.6213 & 0.0475 & 0.0000 & 0.184 & 3.33 \\
\texttt{seed43-Modality-NoVision-Convnext} & 0.3834 & 0.4857 & 0.6192 & 0.0381 & 0.0000 & 0.194 & 3.33 \\
\texttt{seed44-Modality-NoVision-Convnext} & 0.3835 & 0.4864 & 0.6193 & 0.0420 & 0.0000 & 0.192 & 3.33 \\
\texttt{seed42-Modality-NoVision-Mobilenet} & 0.3829 & 0.4847 & 0.6188 & 0.0284 & 0.0000 & 0.199 & 3.33 \\
\texttt{seed43-Modality-NoVision-Mobilenet} & 0.3851 & 0.4877 & 0.6206 & 0.0459 & 0.0000 & 0.180 & 3.33 \\
\texttt{seed44-Modality-NoVision-Mobilenet} & 0.3850 & 0.4879 & 0.6205 & 0.0450 & 0.0000 & 0.193 & 3.33 \\
\texttt{seed42-Modality-NoVision-Resnet} & 0.3832 & 0.4855 & 0.6190 & 0.0384 & 0.0000 & 0.179 & 3.33 \\
\texttt{seed43-Modality-NoVision-Resnet} & 0.3828 & 0.4853 & 0.6187 & 0.0373 & 0.0000 & 0.187 & 3.33 \\
\texttt{seed44-Modality-NoVision-Resnet} & 0.3839 & 0.4861 & 0.6196 & 0.0395 & 0.0000 & 0.192 & 3.33 \\
\texttt{seed42-Modality-NoVision-Simclr} & 0.3835 & 0.4857 & 0.6193 & 0.0380 & 0.0000 & 0.192 & 3.33 \\
\texttt{seed43-Modality-NoVision-Simclr} & 0.3840 & 0.4865 & 0.6197 & 0.0323 & 0.0000 & 0.183 & 3.33 \\
\texttt{seed44-Modality-NoVision-Simclr} & 0.3836 & 0.4860 & 0.6194 & 0.0402 & 0.0000 & 0.190 & 3.33 \\
\texttt{seed42-Modality-NoVision-Swin} & 0.3833 & 0.4854 & 0.6191 & 0.0364 & 0.0000 & 0.190 & 3.33 \\
\texttt{seed43-Modality-NoVision-Swin} & 0.3839 & 0.4865 & 0.6196 & 0.0399 & 0.0000 & 0.184 & 3.33 \\
\texttt{seed44-Modality-NoVision-Swin} & 0.3837 & 0.4861 & 0.6195 & 0.0402 & 0.0000 & 0.188 & 3.33 \\
\texttt{seed42-Modality-NoVision-Tinycnn} & 0.3851 & 0.4876 & 0.6205 & 0.0447 & 0.0000 & 0.186 & 3.33 \\
\texttt{seed43-Modality-NoVision-Tinycnn} & 0.3837 & 0.4858 & 0.6194 & 0.0374 & 0.0000 & 0.189 & 3.33 \\
\texttt{seed44-Modality-NoVision-Tinycnn} & 0.3837 & 0.4861 & 0.6194 & 0.0395 & 0.0000 & 0.207 & 3.33 \\
\texttt{seed42-Modality-StateOnly-Convnext} & 0.8626 & 0.7202 & 0.9288 & 0.0878 & 0.0022 & 0.166 & 3.18 \\
\texttt{seed43-Modality-StateOnly-Convnext} & 0.8626 & 0.7190 & 0.9288 & 0.0882 & 0.0023 & 0.178 & 3.18 \\
\texttt{seed44-Modality-StateOnly-Convnext} & 0.8627 & 0.7179 & 0.9288 & 0.0885 & 0.0023 & 0.176 & 3.18 \\
\texttt{seed42-Modality-StateOnly-Mobilenet} & 0.8627 & 0.7177 & 0.9288 & 0.0886 & 0.0023 & 0.154 & 3.18 \\
\texttt{seed43-Modality-StateOnly-Mobilenet} & 0.8626 & 0.7208 & 0.9288 & 0.0876 & 0.0022 & 0.182 & 3.18 \\
\texttt{seed44-Modality-StateOnly-Mobilenet} & 0.8626 & 0.7180 & 0.9288 & 0.0885 & 0.0023 & 0.176 & 3.18 \\
\texttt{seed42-Modality-StateOnly-Resnet} & 0.8627 & 0.7175 & 0.9288 & 0.0887 & 0.0023 & 0.175 & 3.18 \\
\texttt{seed43-Modality-StateOnly-Resnet} & 0.8628 & 0.7215 & 0.9289 & 0.0875 & 0.0022 & 0.185 & 3.18 \\
\texttt{seed44-Modality-StateOnly-Resnet} & 0.8627 & 0.7177 & 0.9288 & 0.0886 & 0.0023 & 0.171 & 3.18 \\
\texttt{seed42-Modality-StateOnly-Simclr} & 0.8628 & 0.7165 & 0.9289 & 0.0890 & 0.0023 & 0.181 & 3.18 \\
\texttt{seed43-Modality-StateOnly-Simclr} & 0.8627 & 0.7190 & 0.9288 & 0.0882 & 0.0023 & 0.184 & 3.18 \\
\texttt{seed44-Modality-StateOnly-Simclr} & 0.8629 & 0.7154 & 0.9289 & 0.0893 & 0.0023 & 0.188 & 3.18 \\
\texttt{seed42-Modality-StateOnly-Swin} & 0.8627 & 0.7176 & 0.9288 & 0.0886 & 0.0023 & 0.180 & 3.18 \\
\texttt{seed43-Modality-StateOnly-Swin} & 0.8627 & 0.7196 & 0.9288 & 0.0880 & 0.0022 & 0.188 & 3.18 \\
\texttt{seed44-Modality-StateOnly-Swin} & 0.8629 & 0.7224 & 0.9289 & 0.0872 & 0.0022 & 0.180 & 3.18 \\
\texttt{seed42-Modality-StateOnly-Tinycnn} & 0.8628 & 0.7215 & 0.9289 & 0.0875 & 0.0022 & 0.180 & 3.18 \\
\texttt{seed43-Modality-StateOnly-Tinycnn} & 0.8633 & 0.7139 & 0.9291 & 0.0899 & 0.0024 & 0.175 & 3.18 \\
\texttt{seed44-Modality-StateOnly-Tinycnn} & 0.8629 & 0.7227 & 0.9289 & 0.0871 & 0.0022 & 0.180 & 3.18 \\
\end{longtable}
\end{landscape}

\section{Computing Environment}
\label{supp:environment}

Table~\ref{tab:supp_environment} reports the hardware and software environment used for the EVLA experiments. The same workstation and software stack were used for the reported ablation runs to keep runtime and inference-time comparisons consistent.

\begin{table}[t]
\centering
\caption{Hardware and software environment used for the EVLA experiments.}
\label{tab:supp_environment}
\begin{tabular}{ll}
\toprule
Component & Configuration \\
\midrule
Operating system & Ubuntu 22.04.5 LTS \\
CPU & Intel Core Ultra 9 275HX \\
Logical CPUs & 24 \\
System memory & 62~GiB RAM \\
GPU & NVIDIA GeForce RTX 5090 GPU \\
GPU memory & 24~GB \\
NVIDIA driver & 580.159.03 \\
Driver-reported CUDA & 13.0 \\
Python & 3.10.12 \\
PyTorch & 2.9.1+cu128 \\
PyTorch CUDA & 12.8 \\
NumPy & 1.26.4 \\
pandas & 2.3.3 \\
scikit-learn & 1.7.2 \\
Pillow & 12.0.0 \\
OpenCV & 4.13.0 \\
\bottomrule
\end{tabular}
\end{table}

\section{Reproducibility Package}
\label{supp:checklist}

A complete archival EVLA package is defined to contain the manuscript and
supplementary sources, dataset metadata and validation summaries,
representative sample records, aggregated and seed-level ablation results,
model-configuration files, train--validation--test split identifiers or
hashes, environment and dependency information, figure- and table-generation
scripts, and the applicable code and dataset license statements.

The source REFIT measurements are publicly available through the repositories
cited in the main manuscript. The generated EVLA collection is a separate
research artifact derived from those records. Because of its scale, the full
generated collection is intended for distribution through a dedicated dataset
host rather than as part of the manuscript source archive. The supplementary
material itself provides the complete reported result tables and the computing
environment required to inspect the experimental claims.


\begin{thebibliography}{33}
\expandafter\ifx\csname natexlab\endcsname\relax\def\natexlab#1{#1}\fi
\providecommand{\url}[1]{\texttt{#1}}
\providecommand{\href}[2]{#2}
\providecommand{\path}[1]{#1}
\providecommand{\DOIprefix}{doi:}
\providecommand{\ArXivprefix}{arXiv:}
\providecommand{\URLprefix}{URL: }
\providecommand{\Pubmedprefix}{pmid:}
\providecommand{\doi}[1]{\href{http://dx.doi.org/#1}{\path{#1}}}
\providecommand{\Pubmed}[1]{\href{pmid:#1}{\path{#1}}}
\providecommand{\bibinfo}[2]{#2}
\ifx\xfnm\relax \def\xfnm[#1]{\unskip,\space#1}\fi
%Type = Inproceedings
\bibitem[{Arai et~al.(2025)Arai, Miwa, Sasaki, Watanabe, Yamaguchi, Aoki and
  Yamamoto}]{arai2025covla}
\bibinfo{author}{Arai, H.}, \bibinfo{author}{Miwa, K.},
  \bibinfo{author}{Sasaki, K.}, \bibinfo{author}{Watanabe, K.},
  \bibinfo{author}{Yamaguchi, Y.}, \bibinfo{author}{Aoki, S.},
  \bibinfo{author}{Yamamoto, I.}, \bibinfo{year}{2025}.
\newblock \bibinfo{title}{{CoVLA}: Comprehensive vision-language-action dataset
  for autonomous driving}, in: \bibinfo{booktitle}{Proceedings of the Winter
  Conference on Applications of Computer Vision ({WACV})}, pp.
  \bibinfo{pages}{1933--1943}.
\newblock \DOIprefix\doi{10.1109/WACV61041.2025.00195}.
%Type = Inproceedings
\bibitem[{Brohan et~al.(2023)Brohan, Brown, Carbajal, Chebotar, Dabis, Finn,
  Gopalakrishnan, Hausman, Herzog, Hsu, Ibarz, Ichter, Irpan, Jackson,
  Jesmonth, Joshi, Julian, Kalashnikov, Kuang, Leal, Lee, Levine, Lu, Malla,
  Manjunath, Mordatch, Nachum, Parada, Peralta, Perez, Pertsch, Quiambao, Rao,
  Ryoo, Salazar, Sanketi, Sayed, Singh, Sontakke, Stone, Tan, Tran, Vanhoucke,
  Vega, Vuong, Xia, Xiao, Xu, Xu, Yu and Zitkovich}]{brohan2023rt1}
\bibinfo{author}{Brohan, A.}, \bibinfo{author}{Brown, N.},
  \bibinfo{author}{Carbajal, J.}, \bibinfo{author}{Chebotar, Y.},
  \bibinfo{author}{Dabis, J.}, \bibinfo{author}{Finn, C.},
  \bibinfo{author}{Gopalakrishnan, K.}, \bibinfo{author}{Hausman, K.},
  \bibinfo{author}{Herzog, A.}, \bibinfo{author}{Hsu, J.},
  \bibinfo{author}{Ibarz, J.}, \bibinfo{author}{Ichter, B.},
  \bibinfo{author}{Irpan, A.}, \bibinfo{author}{Jackson, T.},
  \bibinfo{author}{Jesmonth, S.}, \bibinfo{author}{Joshi, N.},
  \bibinfo{author}{Julian, R.}, \bibinfo{author}{Kalashnikov, D.},
  \bibinfo{author}{Kuang, Y.}, \bibinfo{author}{Leal, I.},
  \bibinfo{author}{Lee, K.H.}, \bibinfo{author}{Levine, S.},
  \bibinfo{author}{Lu, Y.}, \bibinfo{author}{Malla, U.},
  \bibinfo{author}{Manjunath, D.}, \bibinfo{author}{Mordatch, I.},
  \bibinfo{author}{Nachum, O.}, \bibinfo{author}{Parada, C.},
  \bibinfo{author}{Peralta, J.}, \bibinfo{author}{Perez, E.},
  \bibinfo{author}{Pertsch, K.}, \bibinfo{author}{Quiambao, J.},
  \bibinfo{author}{Rao, K.}, \bibinfo{author}{Ryoo, M.S.},
  \bibinfo{author}{Salazar, G.}, \bibinfo{author}{Sanketi, P.R.},
  \bibinfo{author}{Sayed, K.}, \bibinfo{author}{Singh, J.},
  \bibinfo{author}{Sontakke, S.}, \bibinfo{author}{Stone, A.},
  \bibinfo{author}{Tan, C.}, \bibinfo{author}{Tran, H.},
  \bibinfo{author}{Vanhoucke, V.}, \bibinfo{author}{Vega, S.},
  \bibinfo{author}{Vuong, Q.H.}, \bibinfo{author}{Xia, F.},
  \bibinfo{author}{Xiao, T.}, \bibinfo{author}{Xu, P.}, \bibinfo{author}{Xu,
  S.}, \bibinfo{author}{Yu, T.}, \bibinfo{author}{Zitkovich, B.},
  \bibinfo{year}{2023}.
\newblock \bibinfo{title}{{RT-1}: Robotics transformer for real-world control
  at scale}, in: \bibinfo{booktitle}{Proceedings of Robotics: Science and
  Systems}, \bibinfo{address}{Daegu, Republic of Korea}.
\newblock \DOIprefix\doi{10.15607/RSS.2023.XIX.025}.
%Type = Article
\bibitem[{Din et~al.(2026)Din, Akram, Saad~Saoud, Rosell and
  Hussain}]{din2026vlafusion}
\bibinfo{author}{Din, M.U.}, \bibinfo{author}{Akram, W.},
  \bibinfo{author}{Saad~Saoud, L.}, \bibinfo{author}{Rosell, J.},
  \bibinfo{author}{Hussain, I.}, \bibinfo{year}{2026}.
\newblock \bibinfo{title}{Multimodal fusion with vision-language-action models
  for robotic manipulation: A systematic review}.
\newblock \bibinfo{journal}{Information Fusion} \bibinfo{volume}{129},
  \bibinfo{pages}{104062}.
\newblock \DOIprefix\doi{10.1016/j.inffus.2025.104062}.
%Type = Article
\bibitem[{Drgo{\v{n}}a et~al.(2020)Drgo{\v{n}}a, Arroyo, Cupeiro~Figueroa,
  Blum, Arendt, Kim, Perarnau~Oll{\'e}, Oravec, Wetter, Vrabie and
  Helsen}]{drgona2020mpcbuildings}
\bibinfo{author}{Drgo{\v{n}}a, J.}, \bibinfo{author}{Arroyo, J.},
  \bibinfo{author}{Cupeiro~Figueroa, I.}, \bibinfo{author}{Blum, D.},
  \bibinfo{author}{Arendt, K.}, \bibinfo{author}{Kim, D.},
  \bibinfo{author}{Perarnau~Oll{\'e}, E.}, \bibinfo{author}{Oravec, J.},
  \bibinfo{author}{Wetter, M.}, \bibinfo{author}{Vrabie, D.L.},
  \bibinfo{author}{Helsen, L.}, \bibinfo{year}{2020}.
\newblock \bibinfo{title}{All you need to know about model predictive control
  for buildings}.
\newblock \bibinfo{journal}{Annual Reviews in Control} \bibinfo{volume}{50},
  \bibinfo{pages}{190--232}.
\newblock \DOIprefix\doi{10.1016/j.arcontrol.2020.09.001}.
%Type = Article
\bibitem[{Du et~al.(2026)Du, Luo, Hu, Zhou and Wen}]{du2026adaptive}
\bibinfo{author}{Du, X.}, \bibinfo{author}{Luo, F.}, \bibinfo{author}{Hu, J.},
  \bibinfo{author}{Zhou, W.}, \bibinfo{author}{Wen, J.}, \bibinfo{year}{2026}.
\newblock \bibinfo{title}{Adaptive home energy management to self-motivated
  user preferences via iterative {LLM}-augmented reinforcement learning}.
\newblock \bibinfo{journal}{Applied Energy} \bibinfo{volume}{415},
  \bibinfo{pages}{127976}.
\newblock \DOIprefix\doi{10.1016/j.apenergy.2026.127976}.
%Type = Article
\bibitem[{El~Makroum et~al.(2026)El~Makroum, Zwickl-Bernhard and
  Kranzl}]{elmakroum2026agentic}
\bibinfo{author}{El~Makroum, R.}, \bibinfo{author}{Zwickl-Bernhard, S.},
  \bibinfo{author}{Kranzl, L.}, \bibinfo{year}{2026}.
\newblock \bibinfo{title}{Agentic {AI} home energy management system: A large
  language model framework for residential load scheduling}.
\newblock \bibinfo{journal}{Results in Engineering} \bibinfo{volume}{29},
  \bibinfo{pages}{109857}.
\newblock \DOIprefix\doi{10.1016/j.rineng.2026.109857}.
%Type = Inproceedings
\bibitem[{Ghosh et~al.(2024)Ghosh, Walke, Pertsch, Black, Mees, Dasari, Hejna,
  Kreiman, Xu, Luo, Tan, Chen, Vuong, Xiao, Sanketi, Sadigh, Finn and
  Levine}]{octo2024}
\bibinfo{author}{Ghosh, D.}, \bibinfo{author}{Walke, H.R.},
  \bibinfo{author}{Pertsch, K.}, \bibinfo{author}{Black, K.},
  \bibinfo{author}{Mees, O.}, \bibinfo{author}{Dasari, S.},
  \bibinfo{author}{Hejna, J.}, \bibinfo{author}{Kreiman, T.},
  \bibinfo{author}{Xu, C.}, \bibinfo{author}{Luo, J.}, \bibinfo{author}{Tan,
  Y.L.}, \bibinfo{author}{Chen, L.Y.}, \bibinfo{author}{Vuong, Q.},
  \bibinfo{author}{Xiao, T.}, \bibinfo{author}{Sanketi, P.R.},
  \bibinfo{author}{Sadigh, D.}, \bibinfo{author}{Finn, C.},
  \bibinfo{author}{Levine, S.}, \bibinfo{year}{2024}.
\newblock \bibinfo{title}{Octo: An open-source generalist robot policy}, in:
  \bibinfo{booktitle}{Proceedings of Robotics: Science and Systems},
  \bibinfo{address}{Delft, Netherlands}.
\newblock \DOIprefix\doi{10.15607/RSS.2024.XX.090}.
%Type = Article
\bibitem[{Hochreiter and Schmidhuber(1997)}]{hochreiter1997lstm}
\bibinfo{author}{Hochreiter, S.}, \bibinfo{author}{Schmidhuber, J.},
  \bibinfo{year}{1997}.
\newblock \bibinfo{title}{Long short-term memory}.
\newblock \bibinfo{journal}{Neural Computation} \bibinfo{volume}{9},
  \bibinfo{pages}{1735--1780}.
\newblock \DOIprefix\doi{10.1162/neco.1997.9.8.1735}.
%Type = Techreport
\bibitem[{{International Energy Agency}(2022)}]{iea2022der}
\bibinfo{author}{{International Energy Agency}}, \bibinfo{year}{2022}.
\newblock \bibinfo{title}{Unlocking the Potential of Distributed Energy
  Resources}.
\newblock \bibinfo{type}{Technical Report}. International Energy Agency.
  \bibinfo{address}{Paris, France}.
\newblock \URLprefix
  \url{https://www.iea.org/reports/unlocking-the-potential-of-distributed-energy-resources}.
%Type = Techreport
\bibitem[{{International Energy Agency}(2024a)}]{iea2024batteries}
\bibinfo{author}{{International Energy Agency}}, \bibinfo{year}{2024}a.
\newblock \bibinfo{title}{Batteries and Secure Energy Transitions}.
\newblock \bibinfo{type}{Technical Report}. International Energy Agency.
  \bibinfo{address}{Paris, France}.
\newblock \URLprefix
  \url{https://www.iea.org/reports/batteries-and-secure-energy-transitions}.
%Type = Techreport
\bibitem[{{International Energy Agency}(2024b)}]{iea2024ev}
\bibinfo{author}{{International Energy Agency}}, \bibinfo{year}{2024}b.
\newblock \bibinfo{title}{Global {EV} Outlook 2024}.
\newblock \bibinfo{type}{Technical Report}. International Energy Agency.
  \bibinfo{address}{Paris, France}.
\newblock \URLprefix \url{https://www.iea.org/reports/global-ev-outlook-2024}.
%Type = Techreport
\bibitem[{{International Energy Agency}(2024c)}]{iea2024renewables}
\bibinfo{author}{{International Energy Agency}}, \bibinfo{year}{2024}c.
\newblock \bibinfo{title}{Renewables 2024}.
\newblock \bibinfo{type}{Technical Report}. International Energy Agency.
  \bibinfo{address}{Paris, France}.
\newblock \URLprefix \url{https://www.iea.org/reports/renewables-2024}.
%Type = Article
\bibitem[{Jung(2026)}]{jung2026hema}
\bibinfo{author}{Jung, W.}, \bibinfo{year}{2026}.
\newblock \bibinfo{title}{Multi-agent home energy management assistant
  ({HEMA})}.
\newblock \bibinfo{journal}{SoftwareX} \bibinfo{volume}{34},
  \bibinfo{pages}{102633}.
\newblock \DOIprefix\doi{10.1016/j.softx.2026.102633}.
%Type = Inproceedings
\bibitem[{Kim et~al.(2025)Kim, Pertsch, Karamcheti, Xiao, Balakrishna, Nair,
  Rafailov, Foster, Sanketi, Vuong, Kollar, Burchfiel, Tedrake, Sadigh, Levine,
  Liang and Finn}]{kim2025openvla}
\bibinfo{author}{Kim, M.J.}, \bibinfo{author}{Pertsch, K.},
  \bibinfo{author}{Karamcheti, S.}, \bibinfo{author}{Xiao, T.},
  \bibinfo{author}{Balakrishna, A.}, \bibinfo{author}{Nair, S.},
  \bibinfo{author}{Rafailov, R.}, \bibinfo{author}{Foster, E.P.},
  \bibinfo{author}{Sanketi, P.R.}, \bibinfo{author}{Vuong, Q.},
  \bibinfo{author}{Kollar, T.}, \bibinfo{author}{Burchfiel, B.},
  \bibinfo{author}{Tedrake, R.}, \bibinfo{author}{Sadigh, D.},
  \bibinfo{author}{Levine, S.}, \bibinfo{author}{Liang, P.},
  \bibinfo{author}{Finn, C.}, \bibinfo{year}{2025}.
\newblock \bibinfo{title}{{OpenVLA}: An open-source vision-language-action
  model}, in: \bibinfo{booktitle}{Proceedings of the 8th Conference on Robot
  Learning}, \bibinfo{publisher}{PMLR}. pp. \bibinfo{pages}{2679--2713}.
\newblock \URLprefix \url{https://proceedings.mlr.press/v270/kim25c.html}.
%Type = Article
\bibitem[{Lato{\'n} et~al.(2024)Lato{\'n}, Grela and
  O{\.z}adowicz}]{laton2024drlhems}
\bibinfo{author}{Lato{\'n}, D.}, \bibinfo{author}{Grela, J.},
  \bibinfo{author}{O{\.z}adowicz, A.}, \bibinfo{year}{2024}.
\newblock \bibinfo{title}{Applications of deep reinforcement learning for home
  energy management systems: A review}.
\newblock \bibinfo{journal}{Energies} \bibinfo{volume}{17},
  \bibinfo{pages}{6420}.
\newblock \DOIprefix\doi{10.3390/en17246420}.
%Type = Inproceedings
\bibitem[{Lin et~al.(2025)Lin, Li, Gao, Yang, Wu, Bai, Lei, Wang and
  Shou}]{lin2025showui}
\bibinfo{author}{Lin, K.Q.}, \bibinfo{author}{Li, L.}, \bibinfo{author}{Gao,
  D.}, \bibinfo{author}{Yang, Z.}, \bibinfo{author}{Wu, S.},
  \bibinfo{author}{Bai, Z.}, \bibinfo{author}{Lei, S.W.},
  \bibinfo{author}{Wang, L.}, \bibinfo{author}{Shou, M.Z.},
  \bibinfo{year}{2025}.
\newblock \bibinfo{title}{{ShowUI}: One vision-language-action model for {GUI}
  visual agent}, in: \bibinfo{booktitle}{Proceedings of the IEEE/CVF Conference
  on Computer Vision and Pattern Recognition}, pp.
  \bibinfo{pages}{19498--19508}.
\newblock \DOIprefix\doi{10.1109/CVPR52734.2025.01816}.
%Type = Inproceedings
\bibitem[{Loshchilov and Hutter(2017)}]{loshchilov2017sgdr}
\bibinfo{author}{Loshchilov, I.}, \bibinfo{author}{Hutter, F.},
  \bibinfo{year}{2017}.
\newblock \bibinfo{title}{{SGDR}: Stochastic gradient descent with warm
  restarts}, in: \bibinfo{booktitle}{International Conference on Learning
  Representations}.
\newblock \URLprefix \url{https://openreview.net/forum?id=Skq89Scxx}.
%Type = Inproceedings
\bibitem[{Loshchilov and Hutter(2019)}]{loshchilov2019adamw}
\bibinfo{author}{Loshchilov, I.}, \bibinfo{author}{Hutter, F.},
  \bibinfo{year}{2019}.
\newblock \bibinfo{title}{Decoupled weight decay regularization}, in:
  \bibinfo{booktitle}{International Conference on Learning Representations}.
\newblock \URLprefix \url{https://openreview.net/forum?id=Bkg6RiCqY7}.
%Type = Article
\bibitem[{Michailidis et~al.(2025)Michailidis, Michailidis, Minelli, Coban and
  Kosmatopoulos}]{michailidis2025mpc}
\bibinfo{author}{Michailidis, P.}, \bibinfo{author}{Michailidis, I.},
  \bibinfo{author}{Minelli, F.}, \bibinfo{author}{Coban, H.H.},
  \bibinfo{author}{Kosmatopoulos, E.}, \bibinfo{year}{2025}.
\newblock \bibinfo{title}{Model predictive control for smart buildings:
  Applications and innovations in energy management}.
\newblock \bibinfo{journal}{Buildings} \bibinfo{volume}{15},
  \bibinfo{pages}{3298}.
\newblock \DOIprefix\doi{10.3390/buildings15183298}.
%Type = Inproceedings
\bibitem[{Michelon et~al.(2025)Michelon, Zhou and
  Morstyn}]{michelon2025llmhems}
\bibinfo{author}{Michelon, F.}, \bibinfo{author}{Zhou, Y.},
  \bibinfo{author}{Morstyn, T.}, \bibinfo{year}{2025}.
\newblock \bibinfo{title}{Large language model interface for home energy
  management systems}, in: \bibinfo{booktitle}{Proceedings of the 16th ACM
  International Conference on Future and Sustainable Energy Systems}, pp.
  \bibinfo{pages}{590--602}.
\newblock \DOIprefix\doi{10.1145/3679240.3734586}.
%Type = Misc
\bibitem[{Murray et~al.(2016)Murray, Stankovic and
  Stankovic}]{murray2016refitcleaned}
\bibinfo{author}{Murray, D.}, \bibinfo{author}{Stankovic, L.},
  \bibinfo{author}{Stankovic, V.}, \bibinfo{year}{2016}.
\newblock \bibinfo{title}{{REFIT}: Electrical load measurements (cleaned)}.
\newblock \bibinfo{howpublished}{University of Strathclyde research data
  repository}.
\newblock \URLprefix
  \url{https://pureportal.strath.ac.uk/en/datasets/refit-electrical-load-measurements-cleaned/},
  \DOIprefix\doi{10.15129/9ab14b0e-19ac-4279-938f-27f643078cec}.
%Type = Article
\bibitem[{Murray et~al.(2017)Murray, Stankovic and Stankovic}]{murray2017refit}
\bibinfo{author}{Murray, D.}, \bibinfo{author}{Stankovic, L.},
  \bibinfo{author}{Stankovic, V.}, \bibinfo{year}{2017}.
\newblock \bibinfo{title}{An electrical load measurements dataset of united
  kingdom households from a two-year longitudinal study}.
\newblock \bibinfo{journal}{Scientific Data} \bibinfo{volume}{4},
  \bibinfo{pages}{160122}.
\newblock \DOIprefix\doi{10.1038/sdata.2016.122}.
%Type = Inproceedings
\bibitem[{{Open X-Embodiment Collaboration}(2024)}]{openx2024}
\bibinfo{author}{{Open X-Embodiment Collaboration}}, \bibinfo{year}{2024}.
\newblock \bibinfo{title}{Open x-embodiment: Robotic learning datasets and
  {RT-X} models}, in: \bibinfo{booktitle}{2024 IEEE International Conference on
  Robotics and Automation}, pp. \bibinfo{pages}{6892--6903}.
\newblock \DOIprefix\doi{10.1109/ICRA57147.2024.10611477}.
%Type = Book
\bibitem[{Pearl(2009)}]{pearl2009causality}
\bibinfo{author}{Pearl, J.}, \bibinfo{year}{2009}.
\newblock \bibinfo{title}{Causality: Models, Reasoning, and Inference}.
\newblock \bibinfo{edition}{2nd} ed., \bibinfo{publisher}{Cambridge University
  Press}, \bibinfo{address}{Cambridge, UK}.
%Type = Book
\bibitem[{Peters et~al.(2017)Peters, Janzing and
  Sch{\"o}lkopf}]{peters2017elements}
\bibinfo{author}{Peters, J.}, \bibinfo{author}{Janzing, D.},
  \bibinfo{author}{Sch{\"o}lkopf, B.}, \bibinfo{year}{2017}.
\newblock \bibinfo{title}{Elements of Causal Inference: Foundations and
  Learning Algorithms}.
\newblock Adaptive Computation and Machine Learning Series,
  \bibinfo{publisher}{The MIT Press}, \bibinfo{address}{Cambridge, MA, USA}.
%Type = Article
\bibitem[{Reihani et~al.(2016)Reihani, Motalleb, Ghorbani and
  Saad~Saoud}]{reihani2016peak}
\bibinfo{author}{Reihani, E.}, \bibinfo{author}{Motalleb, M.},
  \bibinfo{author}{Ghorbani, R.}, \bibinfo{author}{Saad~Saoud, L.},
  \bibinfo{year}{2016}.
\newblock \bibinfo{title}{Load peak shaving and power smoothing of a
  distribution grid with high renewable energy penetration}.
\newblock \bibinfo{journal}{Renewable Energy} \bibinfo{volume}{86},
  \bibinfo{pages}{1372--1379}.
\newblock \DOIprefix\doi{10.1016/j.renene.2015.09.050}.
%Type = Article
\bibitem[{Saad~Saoud et~al.(2022)Saad~Saoud, Al-Marzouqi and
  Hussein}]{saoud2022household}
\bibinfo{author}{Saad~Saoud, L.}, \bibinfo{author}{Al-Marzouqi, H.},
  \bibinfo{author}{Hussein, R.}, \bibinfo{year}{2022}.
\newblock \bibinfo{title}{Household energy consumption prediction using the
  stationary wavelet transform and transformers}.
\newblock \bibinfo{journal}{IEEE Access} \bibinfo{volume}{10},
  \bibinfo{pages}{5171--5183}.
\newblock \DOIprefix\doi{10.1109/ACCESS.2022.3140818}.
%Type = Article
\bibitem[{Saad~Saoud and Ghorbani(2020)}]{saoud2020octonion}
\bibinfo{author}{Saad~Saoud, L.}, \bibinfo{author}{Ghorbani, R.},
  \bibinfo{year}{2020}.
\newblock \bibinfo{title}{Metacognitive octonion-valued neural networks as they
  relate to time series analysis}.
\newblock \bibinfo{journal}{IEEE Transactions on Neural Networks and Learning
  Systems} \bibinfo{volume}{31}, \bibinfo{pages}{539--548}.
\newblock \DOIprefix\doi{10.1109/TNNLS.2019.2905643}.
%Type = Article
\bibitem[{Saad~Saoud et~al.(2018)Saad~Saoud, Rahmoune, Tourtchine and
  Baddari}]{saoud2018solar}
\bibinfo{author}{Saad~Saoud, L.}, \bibinfo{author}{Rahmoune, F.},
  \bibinfo{author}{Tourtchine, V.}, \bibinfo{author}{Baddari, K.},
  \bibinfo{year}{2018}.
\newblock \bibinfo{title}{A novel method to forecast 24 h of global solar
  irradiation}.
\newblock \bibinfo{journal}{Energy Systems} \bibinfo{volume}{9},
  \bibinfo{pages}{171--193}.
\newblock \DOIprefix\doi{10.1007/s12667-016-0218-4}.
%Type = Inproceedings
\bibitem[{Walke et~al.(2023)Walke, Black, Zhao, Vuong, Zheng, Hansen-Estruch,
  He, Myers, Kim, Du, Lee, Fang, Finn and Levine}]{walke2023bridgedata}
\bibinfo{author}{Walke, H.R.}, \bibinfo{author}{Black, K.},
  \bibinfo{author}{Zhao, T.Z.}, \bibinfo{author}{Vuong, Q.},
  \bibinfo{author}{Zheng, C.}, \bibinfo{author}{Hansen-Estruch, P.},
  \bibinfo{author}{He, A.W.}, \bibinfo{author}{Myers, V.},
  \bibinfo{author}{Kim, M.J.}, \bibinfo{author}{Du, M.}, \bibinfo{author}{Lee,
  A.}, \bibinfo{author}{Fang, K.}, \bibinfo{author}{Finn, C.},
  \bibinfo{author}{Levine, S.}, \bibinfo{year}{2023}.
\newblock \bibinfo{title}{{BridgeData V2}: A dataset for robot learning at
  scale}, in: \bibinfo{booktitle}{Proceedings of the 7th Conference on Robot
  Learning}, \bibinfo{publisher}{PMLR}. pp. \bibinfo{pages}{1723--1736}.
\newblock \URLprefix \url{https://proceedings.mlr.press/v229/walke23a.html}.
%Type = Article
\bibitem[{Wang et~al.(2025)Wang, Wang, Huang, Zhu, Arroyo and
  Li}]{wang2025safedrl}
\bibinfo{author}{Wang, X.}, \bibinfo{author}{Wang, P.}, \bibinfo{author}{Huang,
  R.}, \bibinfo{author}{Zhu, X.}, \bibinfo{author}{Arroyo, J.},
  \bibinfo{author}{Li, N.}, \bibinfo{year}{2025}.
\newblock \bibinfo{title}{Safe deep reinforcement learning for building energy
  management}.
\newblock \bibinfo{journal}{Applied Energy} \bibinfo{volume}{377},
  \bibinfo{pages}{124328}.
\newblock \DOIprefix\doi{10.1016/j.apenergy.2024.124328}.
%Type = Article
\bibitem[{Yu et~al.(2020)Yu, Xie, Xie, Zou, Zhang, Sun, Zhang, Zhang and
  Jiang}]{yu2019drlsmarthome}
\bibinfo{author}{Yu, L.}, \bibinfo{author}{Xie, W.}, \bibinfo{author}{Xie, D.},
  \bibinfo{author}{Zou, Y.}, \bibinfo{author}{Zhang, D.}, \bibinfo{author}{Sun,
  Z.}, \bibinfo{author}{Zhang, L.}, \bibinfo{author}{Zhang, Y.},
  \bibinfo{author}{Jiang, T.}, \bibinfo{year}{2020}.
\newblock \bibinfo{title}{Deep reinforcement learning for smart home energy
  management}.
\newblock \bibinfo{journal}{IEEE Internet of Things Journal}
  \bibinfo{volume}{7}, \bibinfo{pages}{2751--2762}.
\newblock \DOIprefix\doi{10.1109/JIOT.2019.2957289}.
%Type = Article
\bibitem[{Zhang et~al.(2026)Zhang, Cheng, Gui, Zhao, Zhao and
  Yan}]{zhang2026llmhem}
\bibinfo{author}{Zhang, X.}, \bibinfo{author}{Cheng, Y.}, \bibinfo{author}{Gui,
  X.}, \bibinfo{author}{Zhao, H.}, \bibinfo{author}{Zhao, J.},
  \bibinfo{author}{Yan, J.}, \bibinfo{year}{2026}.
\newblock \bibinfo{title}{Large language model-enhanced home energy management
  with dynamic user preference elicitation and hierarchical data-sharing}.
\newblock \bibinfo{journal}{Applied Energy} \bibinfo{volume}{410},
  \bibinfo{pages}{127540}.
\newblock \DOIprefix\doi{10.1016/j.apenergy.2026.127540}.

\end{thebibliography}
\end{document}